\documentclass[11pt]{article}
\pdfoutput=1
\usepackage{authblk}
\usepackage{booktabs}
\usepackage{multirow}

\usepackage{wrapfig,lipsum}

\usepackage[dvipsnames]{xcolor}

\def\beq{\begin{equation} }\def\eeq{\end{equation} }\def\1{\mathbf{1}}

\newcommand{\hanze}{}
\newcommand{\HANZE}{}

\usepackage[framemethod=default]{mdframed}
\usepackage{caption}
\usepackage{indentfirst}
\usepackage{amsmath}

\usepackage{amssymb,booktabs}
\usepackage{mathtools}
\usepackage{amsthm,amsfonts}
\usepackage{multicol}
\usepackage{makecell}
\usepackage{algorithm}
\usepackage{array}
\usepackage{algorithmic}
\usepackage[utf8]{inputenc}
\usepackage[colorlinks,
            linkcolor=red,
            anchorcolor=blue,
            citecolor=blue
            ]{hyperref}

\def\ImportFromMnSymbol#1{%
  \DeclareFontFamily{U} {MnSymbol#1}{}
  \DeclareFontShape{U}{MnSymbol#1}{m}{n}{
   <-6> MnSymbol#15
   <6-7> MnSymbol#16
   <7-8> MnSymbol#17
   <8-9> MnSymbol#18
   <9-10> MnSymbol#19
   <10-12> MnSymbol#110
   <12-> MnSymbol#112}{}
  \DeclareFontShape{U}{MnSymbol#1}{b}{n}{
   <-6> MnSymbol#1-Bold5
   <6-7> MnSymbol#1-Bold6
   <7-8> MnSymbol#1-Bold7
   <8-9> MnSymbol#1-Bold8
   <9-10> MnSymbol#1-Bold9
   <10-12> MnSymbol#1-Bold10
   <12-> MnSymbol#1-Bold12}{}
  \DeclareSymbolFont{MnSy#1} {U} {MnSymbol#1}{m}{n}
}
\newcommand\DeclareMnSymbol[4]{\DeclareMathSymbol{#1}{#2}{MnSy#3}{#4}}
\ImportFromMnSymbol{C}

\DeclareMnSymbol{\smalltriangleright}{\mathbin}{C}{72}
\DeclareMnSymbol{\smalltriangleup}{\mathbin}{C}{73}
\DeclareMnSymbol{\smalltriangleleft}{\mathbin}{C}{74}
\DeclareMnSymbol{\smalltriangledown}{\mathbin}{C}{75}
\DeclareMnSymbol{\filledtriangleright}{\mathbin}{C}{76}
\DeclareMnSymbol{\filledtriangleup}{\mathbin}{C}{77}
\DeclareMnSymbol{\filledtriangleleft}{\mathbin}{C}{78}
\DeclareMnSymbol{\filledtriangledown}{\mathbin}{C}{79}
\DeclareMnSymbol{\smallsquare}{\mathbin}{C}{104}
\DeclareMnSymbol{\filledsquare}{\mathbin}{C}{105}
\DeclareMnSymbol{\smalldiamond}{\mathbin}{C}{108}
\DeclareMnSymbol{\filleddiamond}{\mathbin}{C}{109}
\DeclareMnSymbol{\smallstar}{\mathbin}{C}{128}
\DeclareMnSymbol{\filledstar}{\mathbin}{C}{129}
\DeclareMnSymbol{\thinstar}{\mathbin}{C}{134}

\usepackage{fullpage}
\numberwithin{equation}{section}

\newtheorem{example}{Example}

\newtheorem{proposition}{Proposition}

\ifx\assumption\undefined
\newtheorem{assumption}{Assumption}
\fi

\usepackage{tikz}
\usetikzlibrary{fit,positioning}
\usetikzlibrary{shapes,decorations,arrows,calc,arrows.meta,fit,positioning}
\tikzset{
	-Latex,auto,node distance =1.2 cm and 1.2 cm,semithick,
	state/.style ={circle, draw, minimum width = 1.2 cm},
	state2/.style ={circle, draw, minimum width = 1.2 cm,fill=gray!70},
	point/.style = {circle, draw, inner sep=0.04cm,fill,node contents={}},
	bidirected/.style={Latex-Latex,dashed},
	el/.style = {inner sep=2pt, align=left, sloped}
}

\usepackage[algo2e]{algorithm2e} 

\SetKwInput{KwInput}{Input}
\SetKwInput{KwOutput}{Output}
\SetKwInput{KwReturn}{Return}

\newcommand{\EE}{\mathbf{E}}

\newcommand{\kl}{{\mathrm{KL}}}

\newcommand{\ie}{\emph{i.e.}}

\newcommand{\ModelName}{\textsc{NVF}}

\usepackage{multirow}
\usepackage{tablefootnote}
\usepackage{colortbl}
\usepackage{hhline}

\usepackage{mathtools}

\newcommand*\samethanks[1][\value{footnote}]{\footnotemark[#1]}

\begin{document}
\title{
Normalizing Flow with Variational Latent Representation
}

\author[$\dag$]{Hanze Dong\thanks{Equal contributions.}}
\author[$\ddag$]{Shizhe Diao\samethanks}
\author[$\dag$]{Weizhong Zhang}
\author[$\dag$,$\ddag$]{Tong Zhang}

\affil[$\dag$]{Department of Mathematics\\ HKUST}
\affil[$\ddag$]{Department of Computer Science and Engineering\\ HKUST}

\date{}
\maketitle

\begin{abstract}
	Normalizing flow (NF) has gained popularity over traditional maximum likelihood based methods due to its strong capability to model complex data distributions. However, the standard approach, which maps the observed data to a normal distribution, has difficulty in handling data distributions with multiple relatively isolated modes. To overcome this issue, we propose a new framework based on variational latent representation to improve the practical performance of NF. The idea is to replace the standard normal latent variable with a more general latent representation, jointly learned via Variational Bayes. For example, by taking the latent representation as a discrete sequence, our framework can learn a Transformer model that generates the latent sequence and an NF model that generates continuous data distribution conditioned on the sequence. The resulting method is significantly more powerful than the standard normalization flow approach for generating data distributions with multiple modes. Extensive experiments have shown the advantages of NF with variational latent representation.\footnote{Our code is available at  \url{https://github.com/hendrydong/normalizing-variational-flow}.}
\end{abstract}

\section{Introduction}\label{sec:intro}

Modeling high dimensional data without supervision is a fundamental and challenging problem in statistical learning. Recently, deep generative modeling provides us possibilities to achieve this goal with a generative function between a known distribution $Z$ and observation data $X$. The main idea is to learn a function such that $G_\theta(Z)\approx X$.
Unfortunately, most of these models, such as GANs \cite{goodfellow2014generative} and VAEs \cite{kingma2013auto}, {cannot provide explicit likelihood and fail to perform density evaluation for downstream statistical procedures due to the absence of $p_\theta(x)$ form.}  
Another approach is the flow-based method, which has explicit likelihood and captures isomorphism between $Z$ and $X$.
In other words, $G_\theta$ in NF is an invertible function such that $G^{-1}_\theta(X)\approx Z$.
Recent progress in normalizing flows (NFs) \cite{dinh2014nice,kingma2018glow,dinh2016density,durkan2019neural} and Neural ODEs \cite{chen2018neural,grathwohl2018ffjord} have shown that neural networks are capable of reproducing invertible dynamical systems. 
{Notably, when $G_\theta$ is invertible, the likelihood $p_\theta(x)$ can be explicitly expressed with the Jacobian function of $G_\theta$. Thus, NF provides us a free-form likelihood parametric modeling framework.}
It is quite natural and elegant to perform maximum likelihood estimation (MLE) to learn the above isomorphism. 
\begin{figure*}[t]
	\centering
	\begin{tabular}{c|c}
		\includegraphics[width=5.2cm,trim=70 70 50 70,clip]{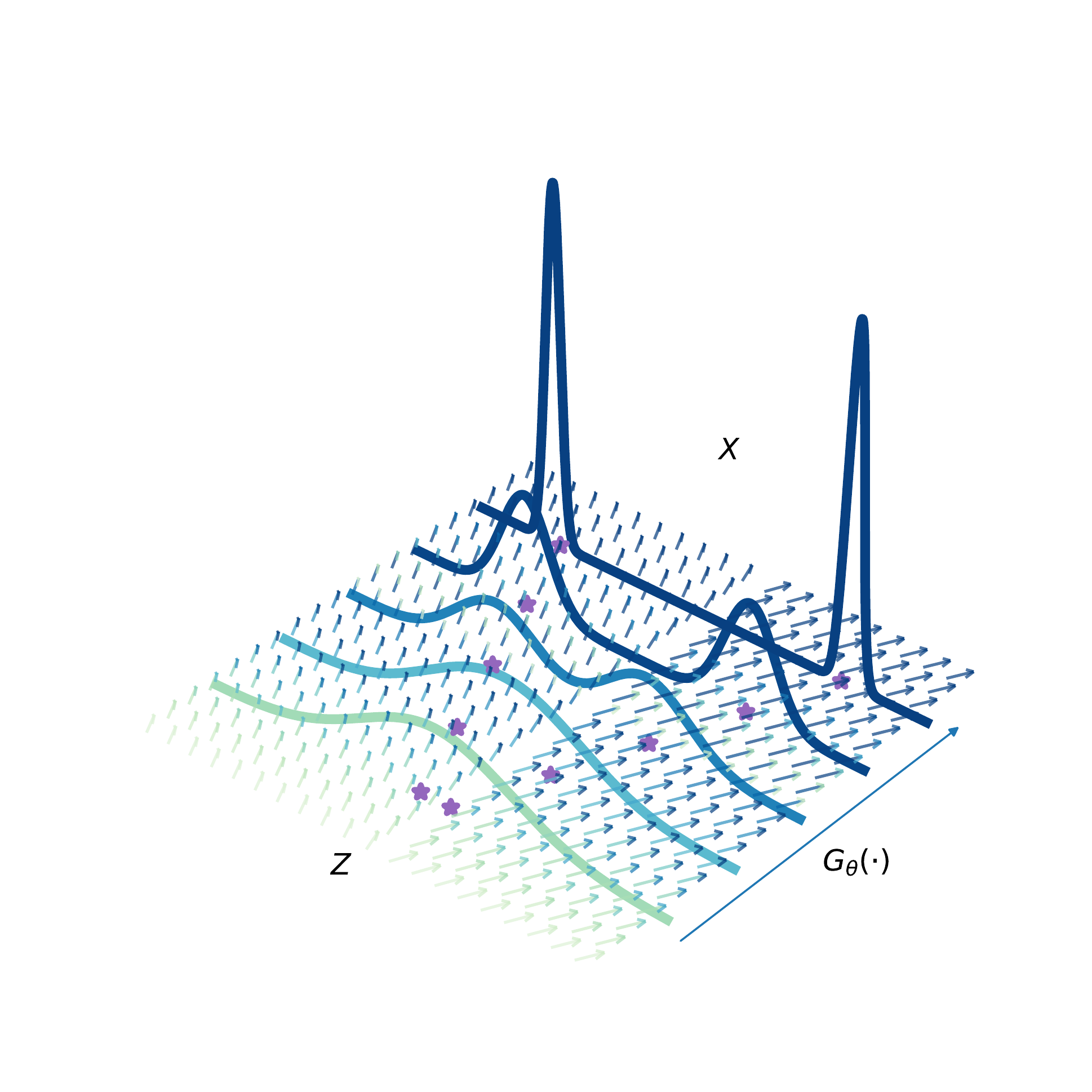}&
		\includegraphics[width=5.2cm,trim=70 70 50 70,clip]{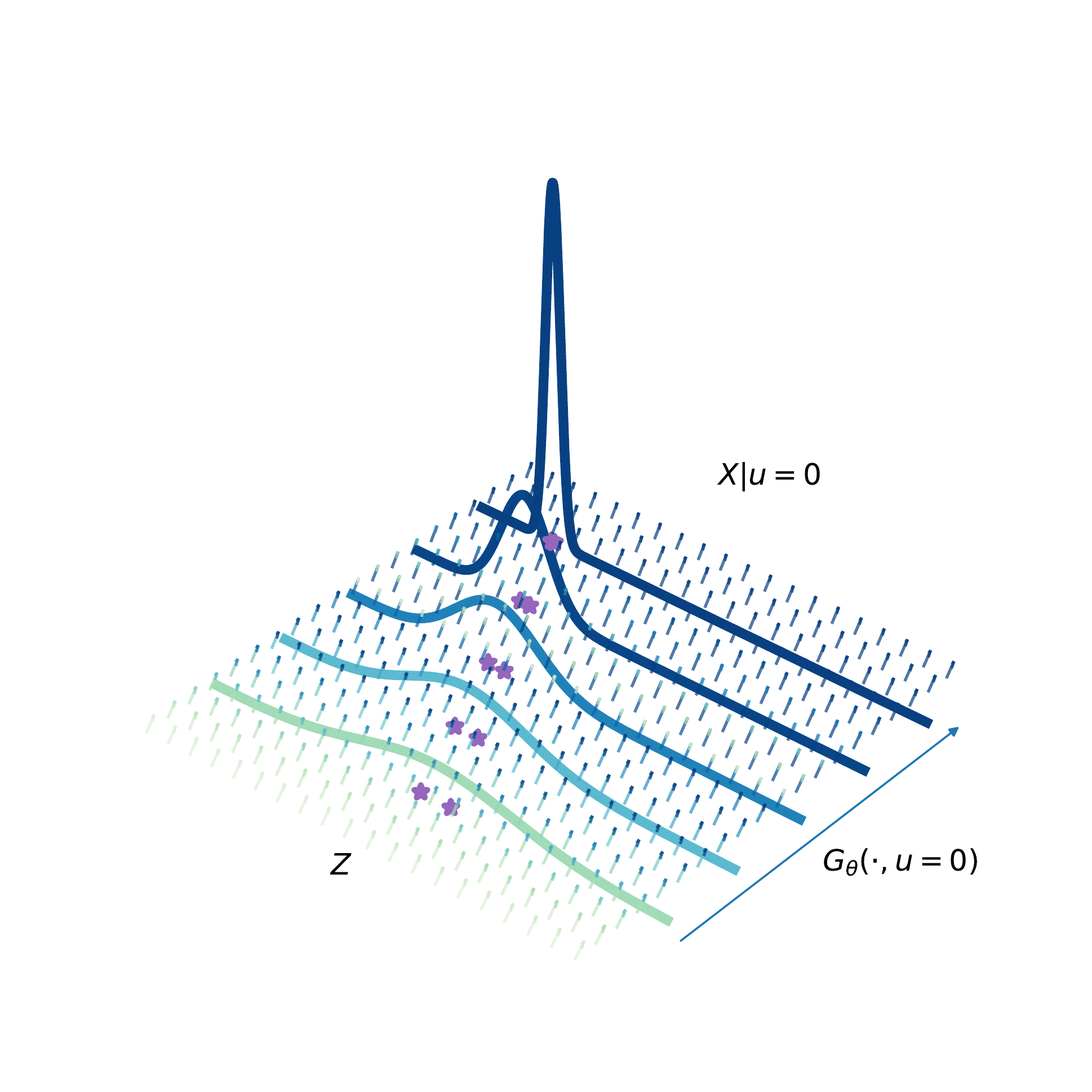}
		\includegraphics[width=5.2cm,trim=70 70 50 70,clip]{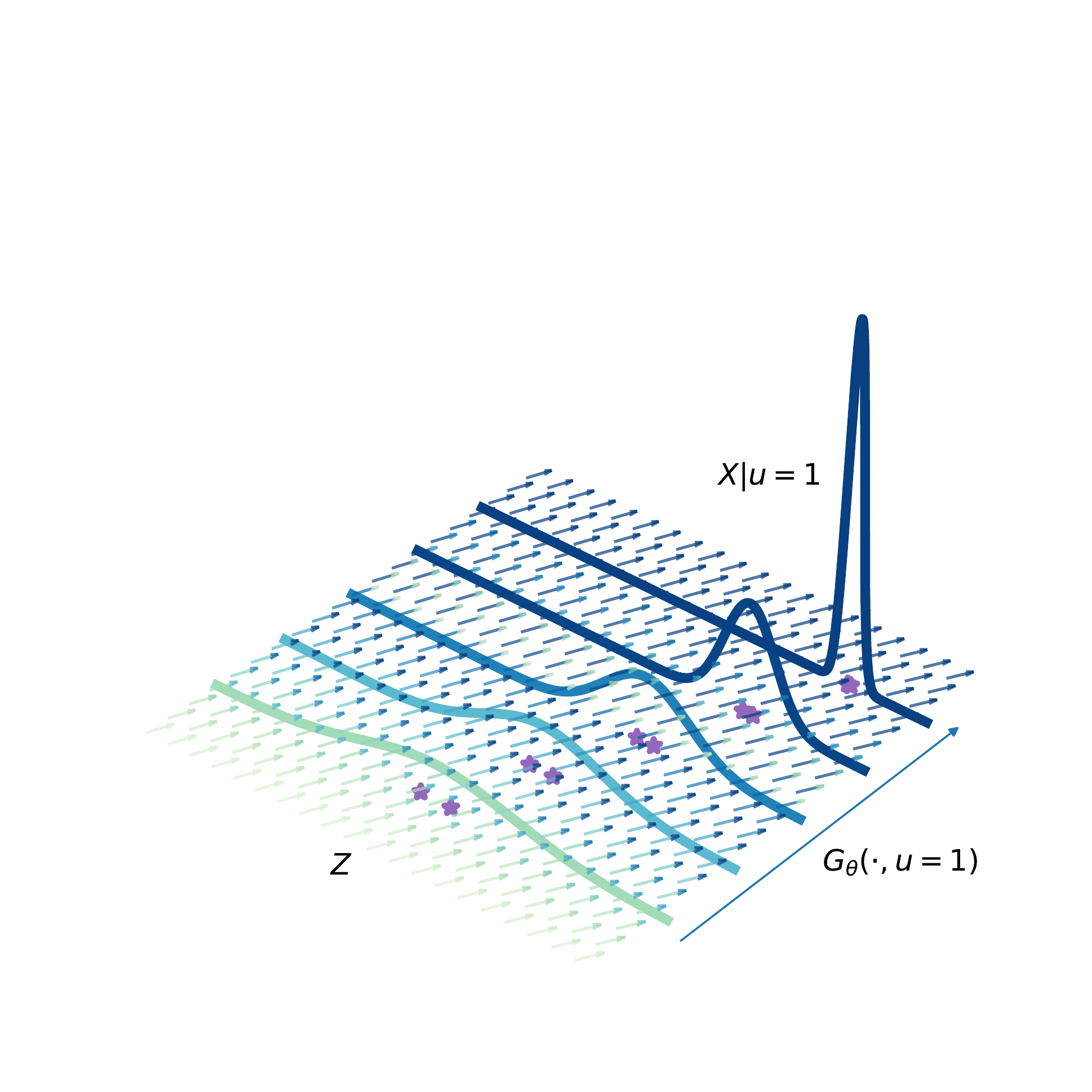}
		\\
		w/o latent factor&with latent factor $u\in\{0,1\}$\\
	\end{tabular}
	\caption{Illustration of necessity of latent variable. The plot of $G_\theta$ contains the full transportation process for better visualization. Purple \textcolor{violet}{$\filledstar$} denote two neighborhood particles in $Z$. NF without $u$ distributes them into different clusters, which makes Jacobian of $G_\theta$ ill-behaved.}
	\label{fig:illustration}
	
\end{figure*}

Intuitively, due to the invertibility, we can regard $G_\theta$ as a transportation function, which conveys $Z$ to $X$.
For simple transportation, such as shift and scale, $G_\theta$ can be easily learned. Thus, NF framework works well for these examples.
Nonetheless, for real data, distributions are often scattered into different clusters (\ie, modes). When these clusters are isolated, the optimal $G$ would be complex, \emph{i.e.}, there are some ``discontinuity" in the function with bad Jacobian behavior.
This makes the $G_\theta$ extremely hard to learn. 
 We can consider an example in Figure \ref{fig:illustration} that $X$ is from a Gaussian mixture model. Obviously, the continuity of transportation function is rather poor near {the center of $Z$}, and Gaussian initialization puts most density on this area.
Thus, learning the invertible dynamics of transportation becomes highly unstable in these discontinuous part. In other words, it is rather difficult to model a scattered distribution by a function from Gaussian.
Details are given in Section \ref{sec:back}.

\hanze{In this paper, to resolve the above limitation of NF, we propose a new framework called Normalizing Variational Flow,  ({\ModelName})}. Specifically, we decompose $X$ into parts, each of which contains only a unimodal, highly concentrated distribution cluster. From this view, we do NOT need to match $Z$ to the full space of $X$, but only find the bijection from $Z$ to each cluster of $X$ separately. Inspired by clustering techniques, we can incorporate a latent variable to denote the representation of each cluster of $X$ and simplify the transportation function with the latent knowledge. The feasibility of the idea is guaranteed by Universal Approximation Theorem of mixture model: 
even a mixture of Gaussian can approximate any smooth density functions \cite{mclachlan2000finite}.
For example, in Figure \ref{fig:illustration}, we divide data samples into two categories $0$ and $1$. By introducing the latent factor $u$, the transportation function  becomes  linear without any discontinuity. 
In real applications, when such latent variables, such as image labels, attributes, are given,  previous studies \cite{salimans2016improved,karras2020analyzing,grathwohl2019your} have demonstrated that leveraging these variables can effectively improve of the generation quality by simplifying $G_\theta$. Beyond supervised labels, for real data, the latent variables exist in a more general form: discrete, continuous, or even sequential latent representations are reasonable candidates and they are usually not accessible.  \HANZE{{Therefore, how to design a proper latent variable $u$ and how to effectively learn it become two crucial problems.  }}

We argue that we should adopt  latent variables which are  more fine-grained than supervised labels. 
The reason is that the number of modes can be far more than that of labels.
Without a proper structure of $u$, the desired transportation function cannot be obtained smoothly.
In general, $u$ should contain global information of each $x$ that makes NF transportation do not need to match the whole population, but one mode of it.  For different data structures and modeling purposes, the design and parameterization of $u$ varies: 
For tabular data or toy data, the key latent variable is the cluster center, we use finite categories as the discrete probability space of $u$;
For complex data, such as images and texts, the cluster center cannot be easily expressed as a single vector, so we introduce a general sequential representation for them. \cite{dosovitskiy2020image} stress that images should be represented as $16\times16$ sequences.

Our approach supports multiple forms of $u$ that are preferred in different settings, which are important in real practice. Specifically, we 
adopt and extend Variational Bayesian (VB) method to jointly learn $G_{\theta}$ and the latent variable $u$. The posterior $\pi_\theta(u|x)$ is our target to assign $u$ for each $x$.
Existing MLE based methods, such as expectation–maximization (EM) algorithm 
\cite{moon1996expectation}, necessitate the explicit form probability measure of $\pi_\theta(u|x)$ in the parametric model. However,  this strategy is intractable due to the traversal requirement of all possible $u$. 
Variational Bayesian methods provide us an alternative to learn another probability
$\nu_\phi(u|x)$ to approximate $\pi_\theta(u|x)$. In our context, this $\nu_\phi(u|x)$ can be parameterized by a neural network encoder, which is flexible about the type of $u$.
By incorporating Variational Bayes, the NF framework can be optimized with evidence lower bound (ELBO). 
This ELBO-based framework mitigates the fatal issue caused by ill-behaved Jacobian matrix in NF.

		\vspace{0.2in}
		\noindent\textbf{Contributions.}
		\begin{itemize}
			\item We point out that conventional NF that
			maps the observed data to a normal distribution has difficulty in handling data distributions with multiple isolated modes;
			\item We propose a framework to optimize NF models based on VB that resolves the aforementioned issue. It employs a variational latent representation to model cluster information, so that a simplified transportation function can be used for each cluster;
			\item Our variational latent framework supports different types of latent factor $u$ that can be customized for different data modeling tasks;
			\item Experiments demonstrated the importance of variational latent representation for NF, and the effectiveness of {\ModelName}.
		\end{itemize}

		\section{Background}\label{sec:back}
		
		\begin{figure*}[t]
			\centering
			\begin{tabular}{cc}
				\includegraphics[width=5.5cm]{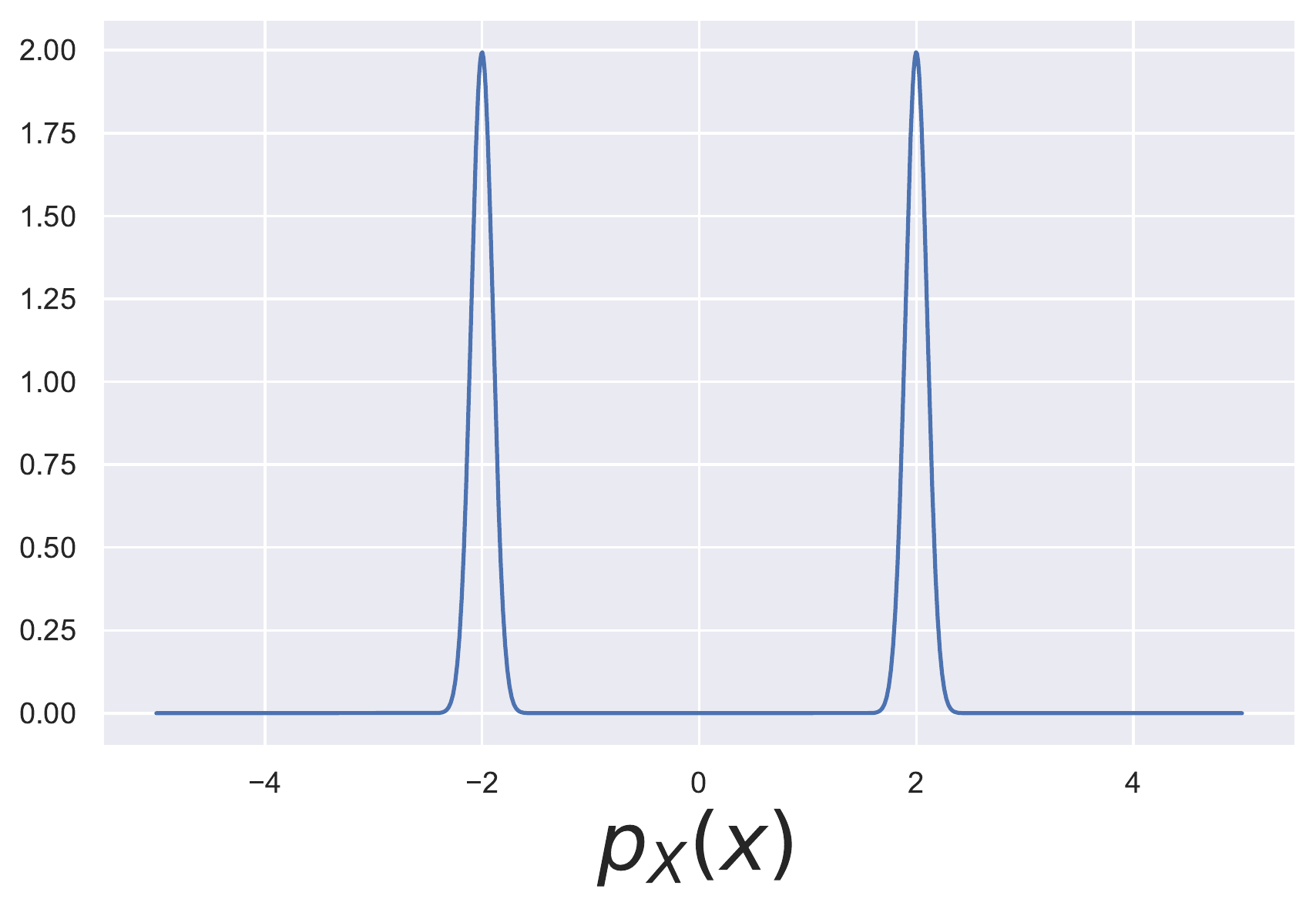}&\includegraphics[width=5.5cm]{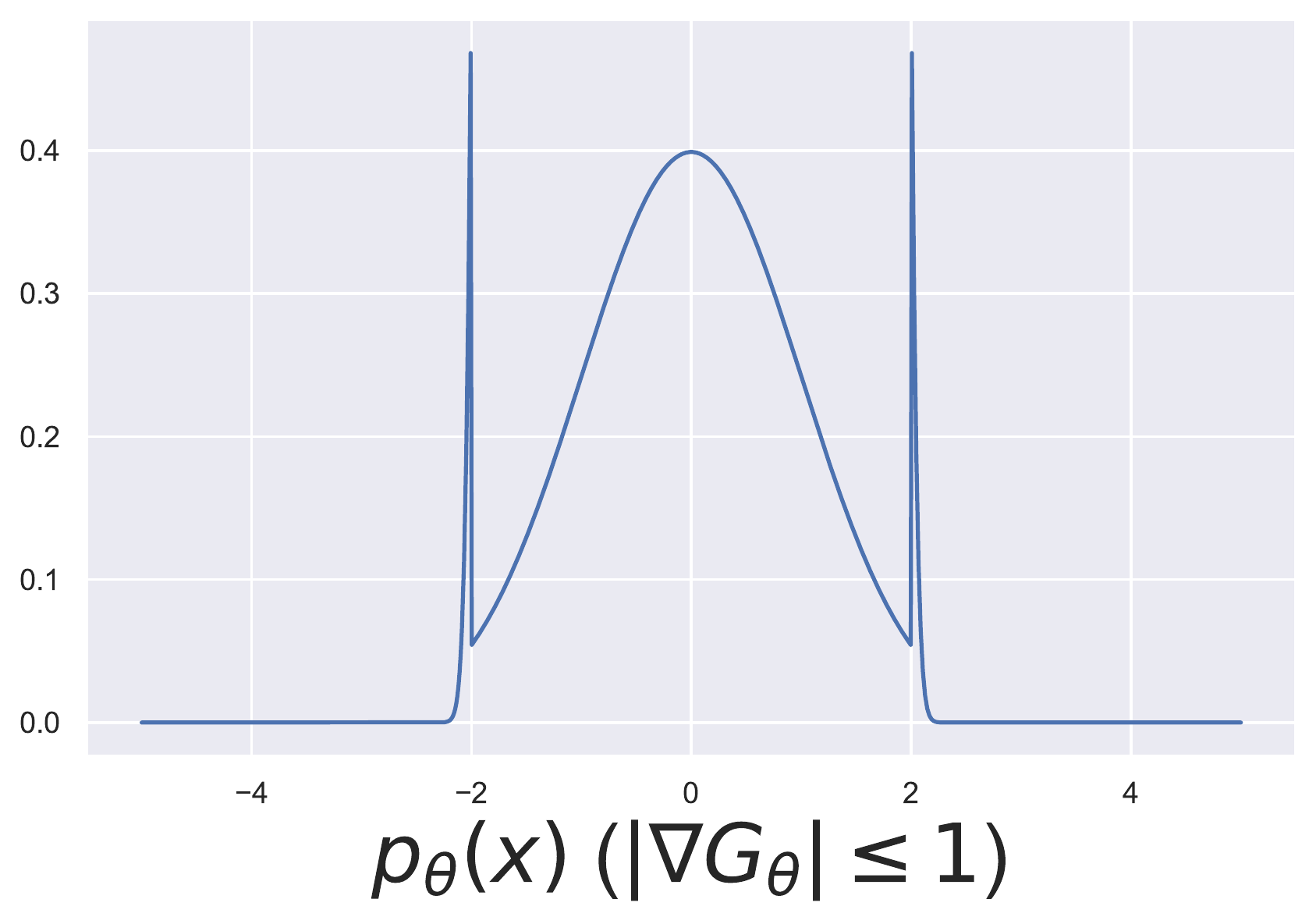}\\\includegraphics[width=5.5cm]{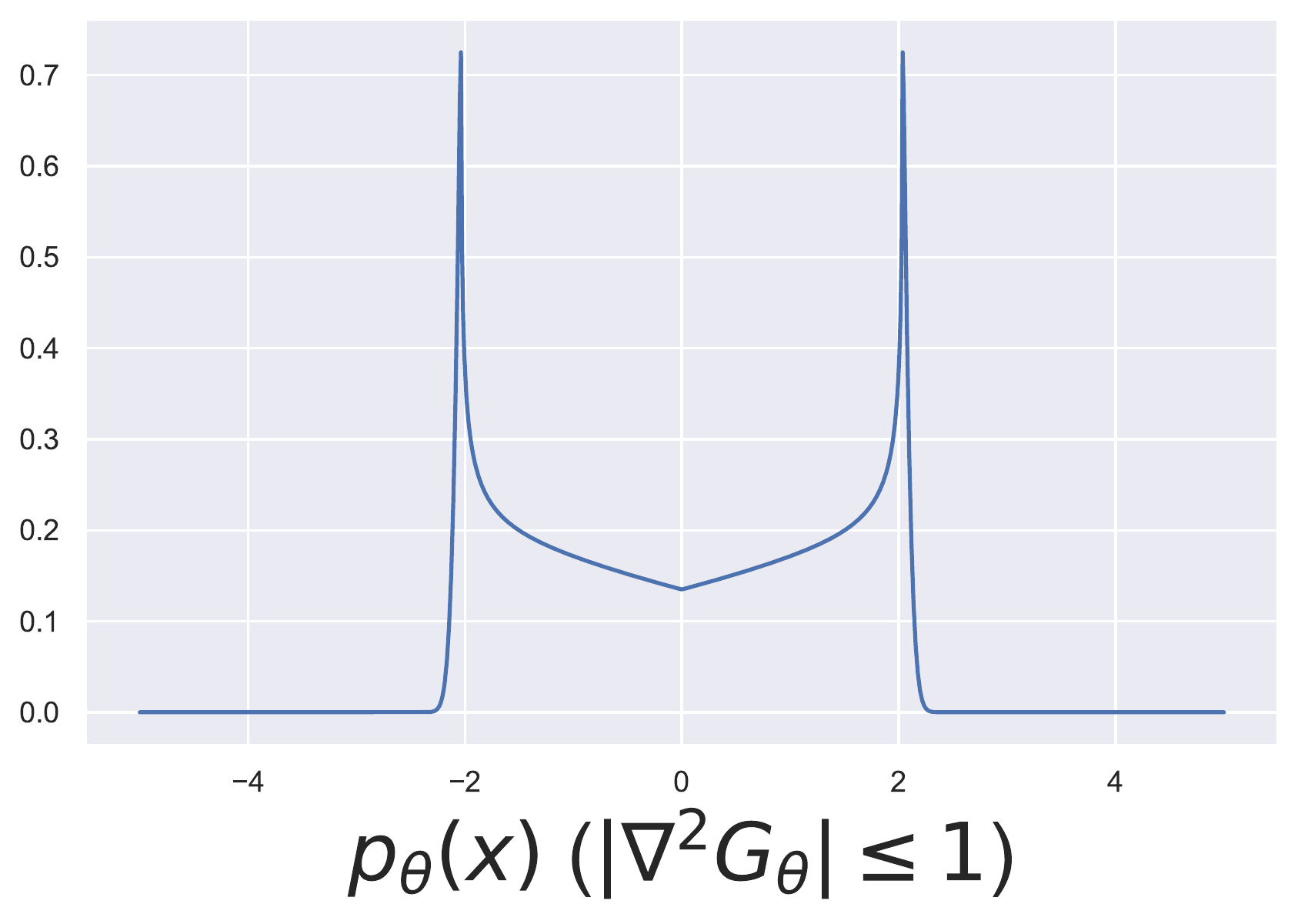}&
				\includegraphics[width=5.5cm]{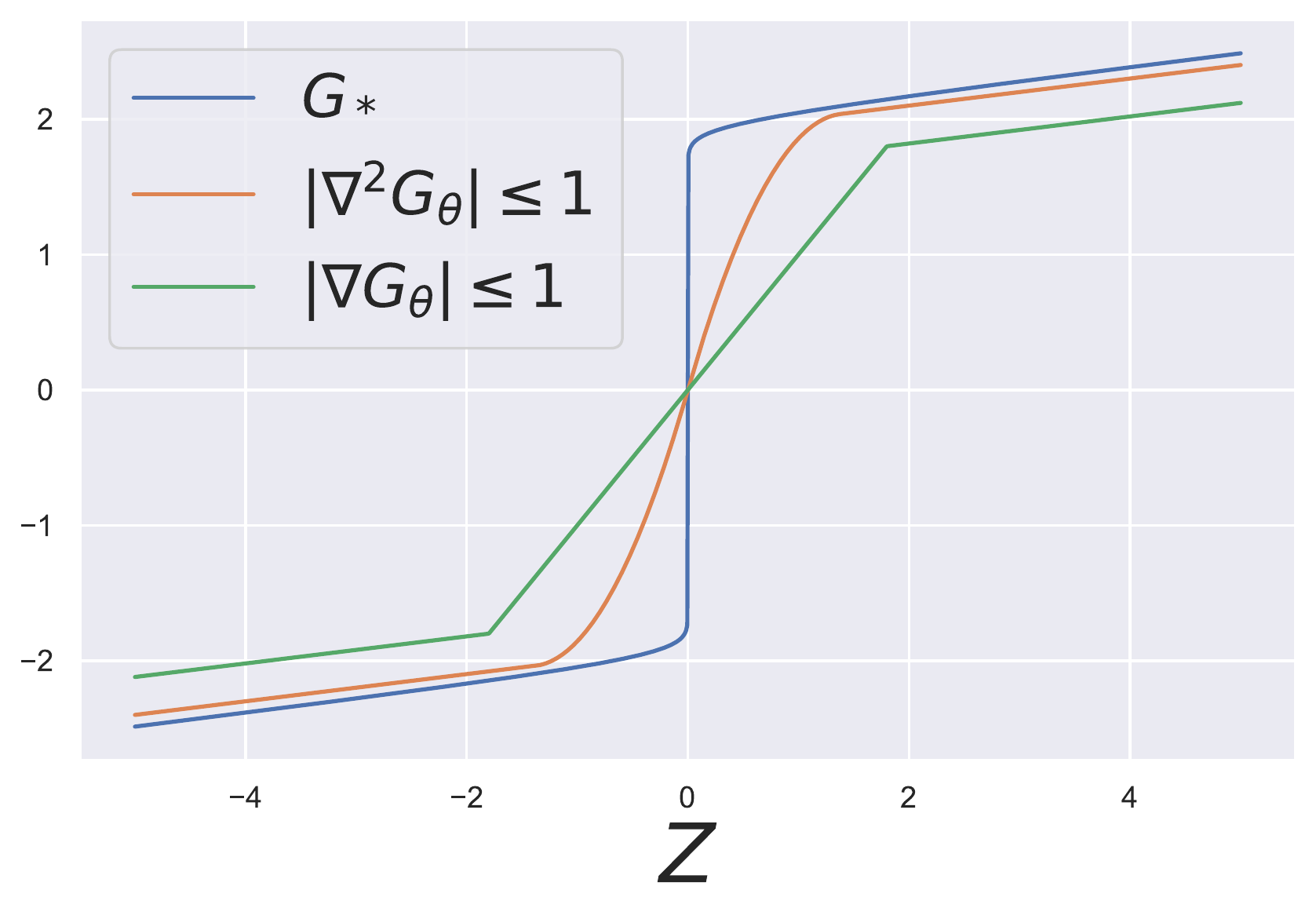}
			\end{tabular}
			\caption{Failure of $G_*$ approximation when $X$ is a mixture of $\mathcal N(\pm 2,0.1^2)$ (MLE-based $\theta$).}
			
			\label{fig:approx_fail}
		\end{figure*}
		
		\begin{figure*}[t]
			\centering
			\begin{tabular}{cc}
				\includegraphics[width=6cm]{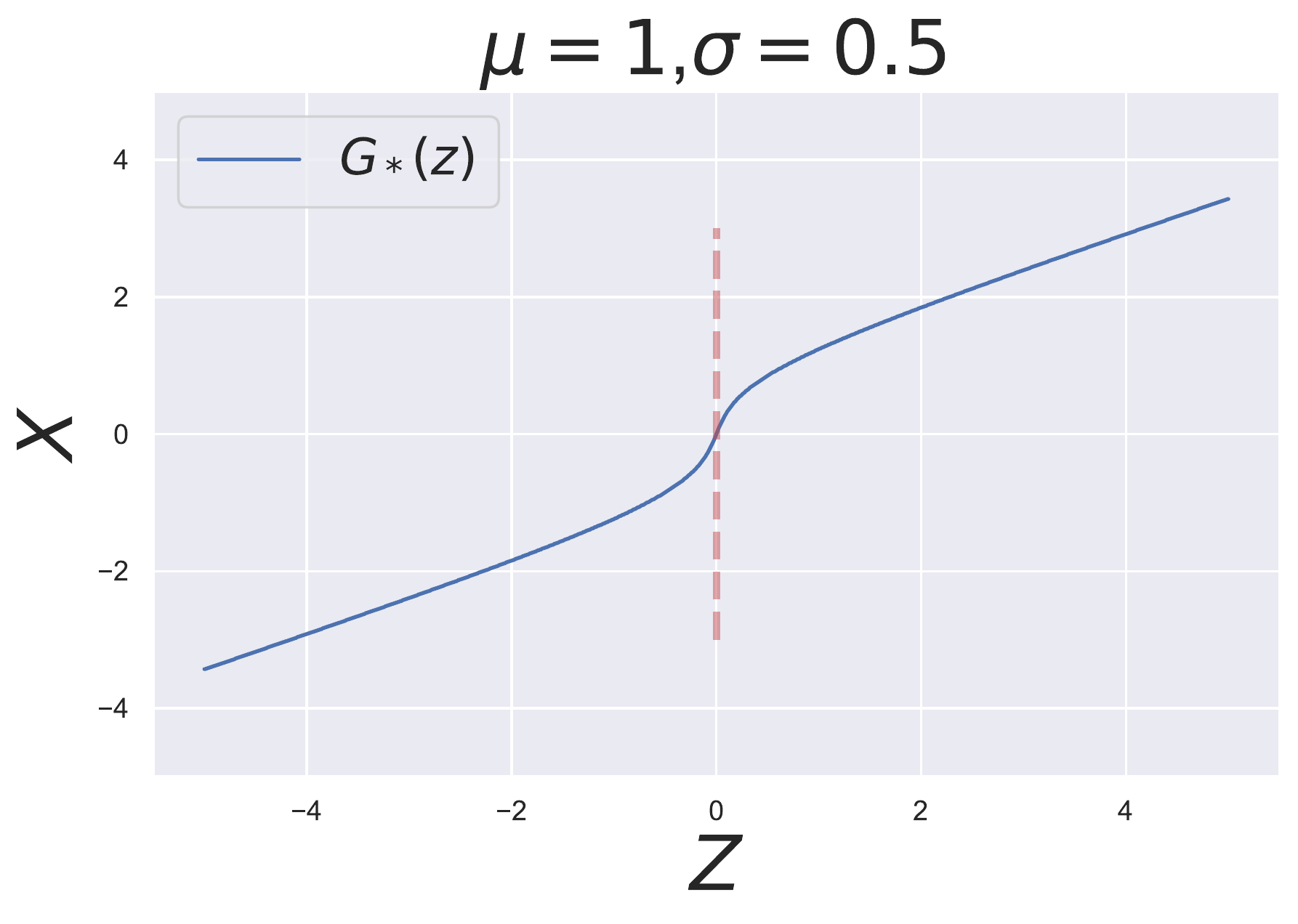}&
				\includegraphics[width=6cm]{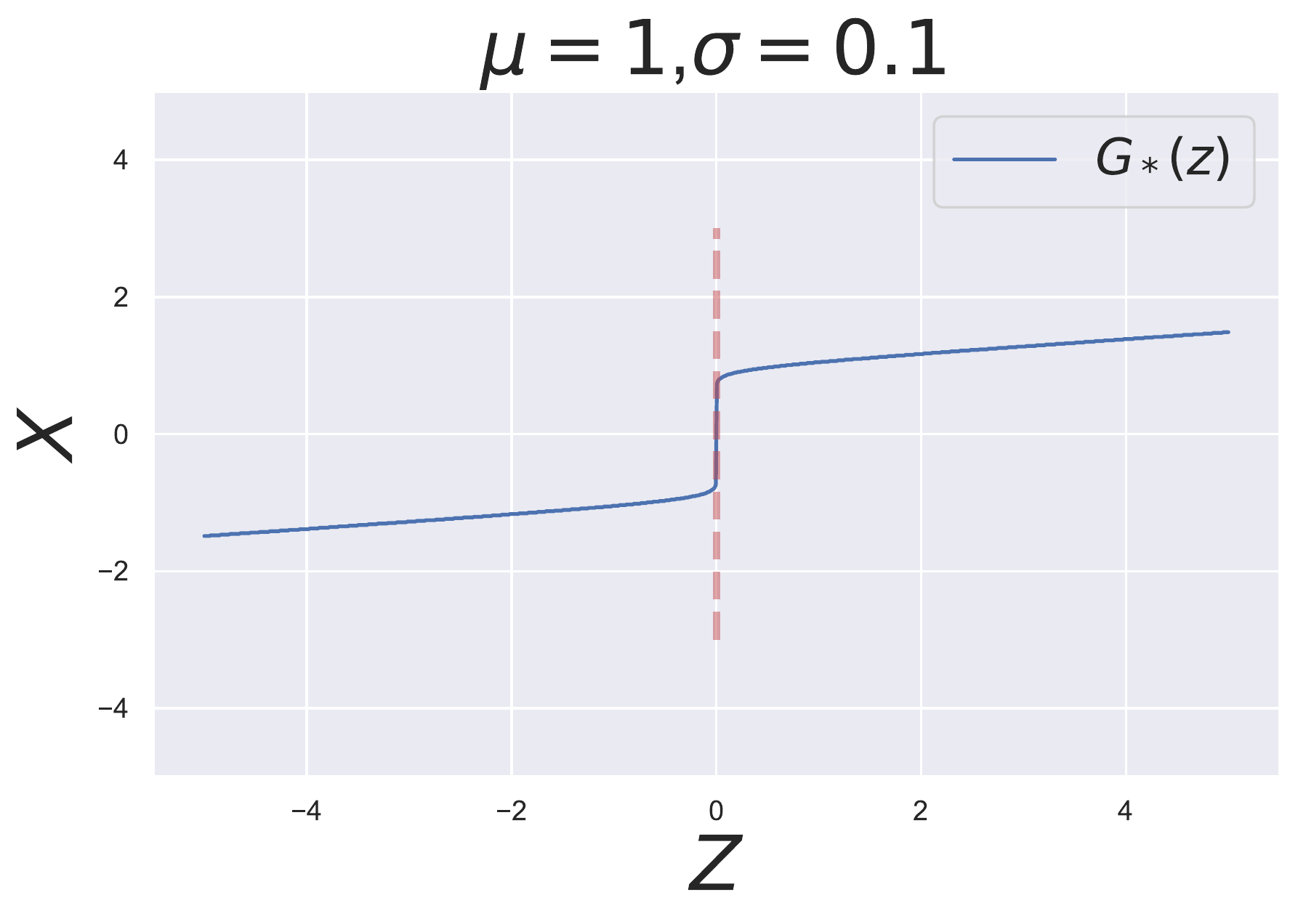}\\
				\includegraphics[width=6cm]{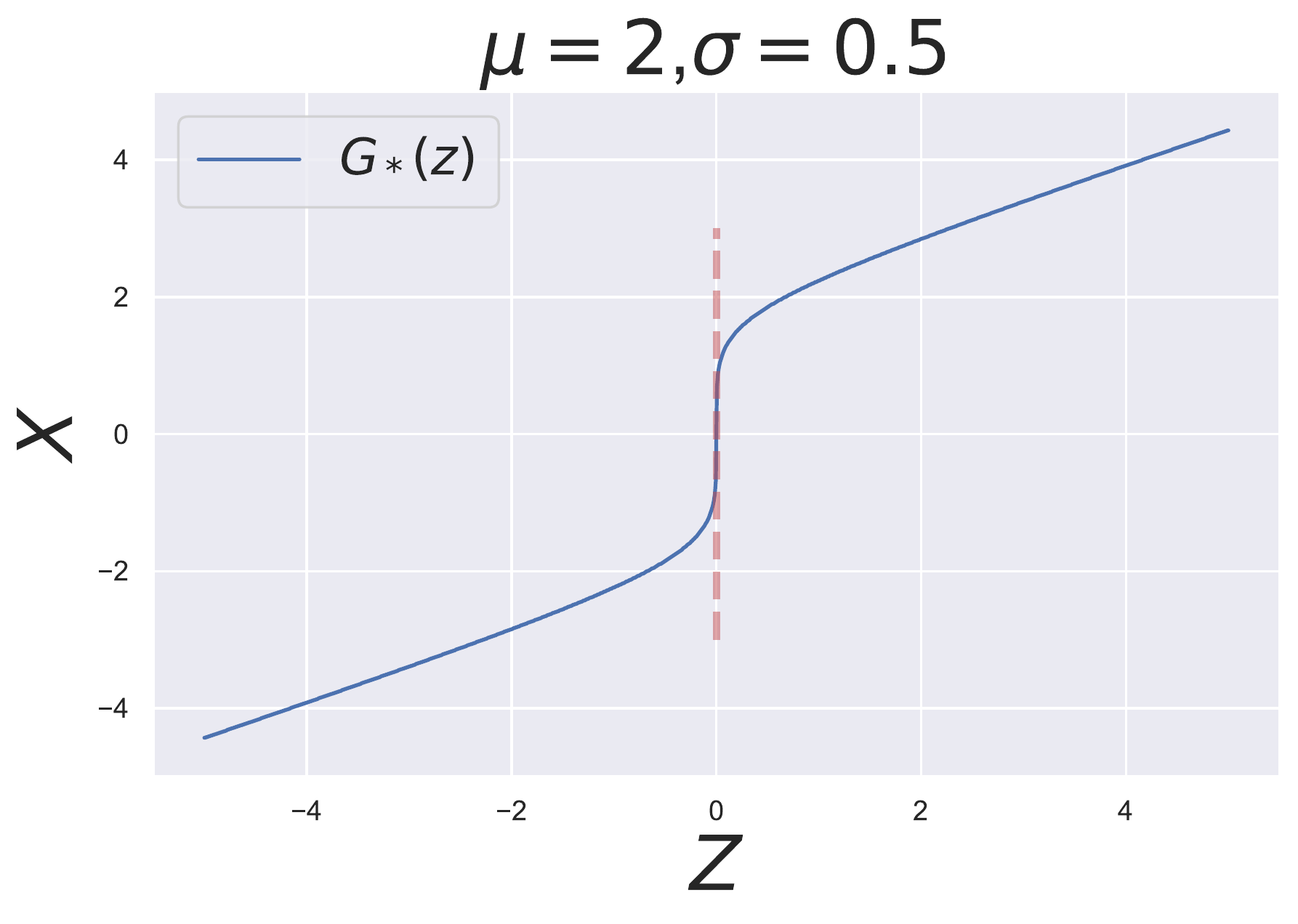}&
				\includegraphics[width=6cm]{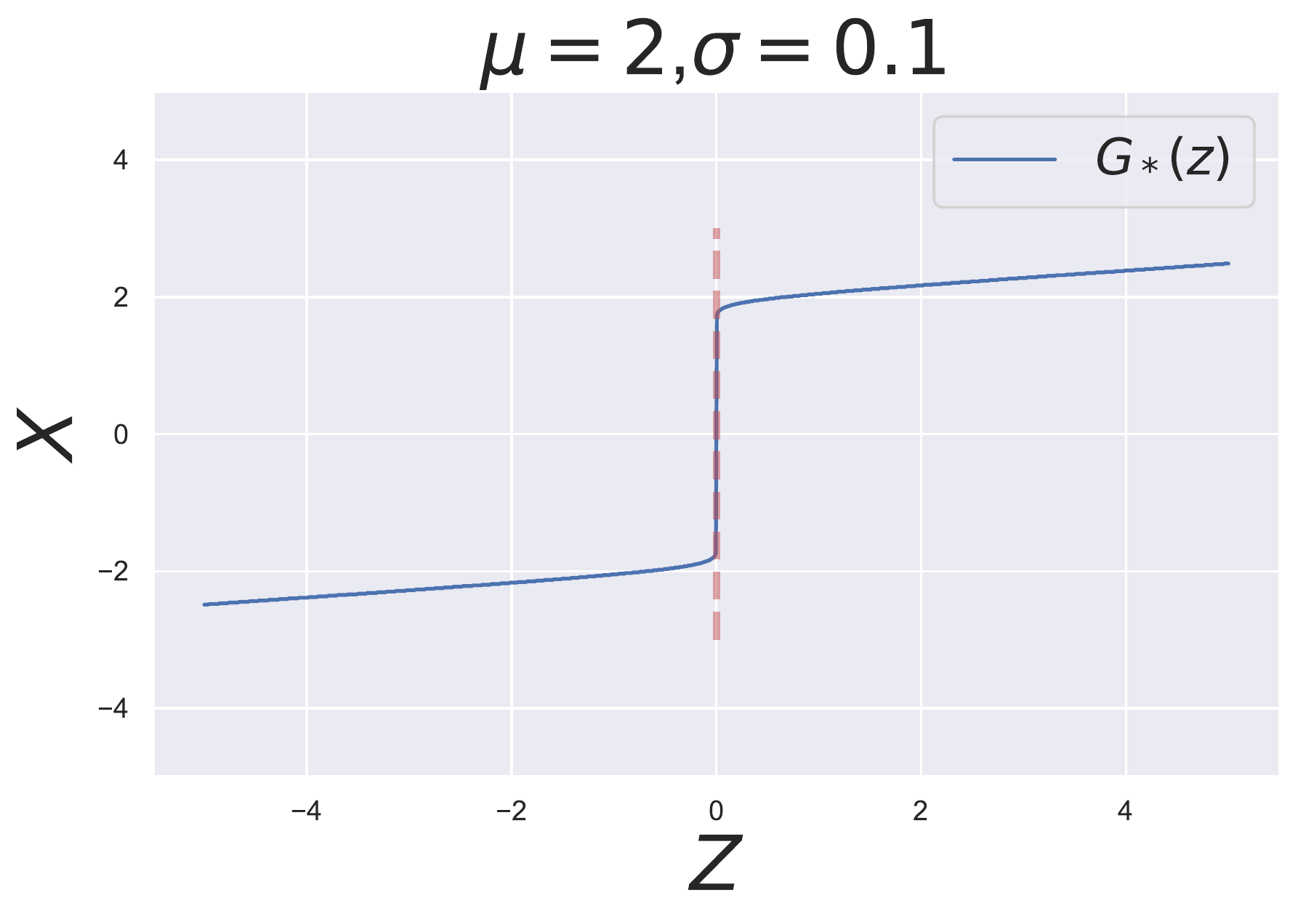}
			\end{tabular}
			\caption{Clustering phenomenon brings ill-behaved Jacobian}
			\label{fig:disconti}
		\end{figure*}
		
		\noindent\textbf{Notations.} Throughout this paper, 
		capital letters denote random variables and lower-case letters denote data samples. $p_\theta$ denotes the density parameterized by $\theta$, $p_X$ and $p_Z$ denote data distribution $X$ and
		a known density of $Z$, respectively. \hanze{{$\nabla G_\theta(x)$ and $\nabla_xG_\theta(x)$ denote the Jacobian of $G$ w.r.t. $x$}}. $\nabla_\theta G_\theta(x)$ is the Jacobian  w.r.t. $\theta$. $|J|$ is the determinant of a matrix $J$. \hanze{ $X$ and $Z$ are continuous random variables.} For latent random variable $U$, it can be continuous, discrete, or sequential. The probability measure induced by $\beta$ and $\phi$ is $\pi_\beta$ and $\nu_\phi$, the Radon–Nikodym derivative \cite{durrett2019probability} of $\pi_\beta$ and $\nu_\phi$ over the measure $\mu_u$ of $U$ is denoted by $p_\beta$ and $p_\phi$, which 
		is the generalized definition for both discrete and continuous random variables.
		$\operatorname{KL}(p\Vert q)$ denotes the Kullback–Leibler (KL) divergence of $p$ from $q$. $\mathcal N(\mu,\Sigma)$ denotes the Gaussian distribution with mean/covariance $\mu$ and $\Sigma$.

		\vspace{.2in}
		\noindent\textbf{Normalizing Flows and Latent Variable.}
		This work considers the dynamics of normalizing flow, which aims to find an invertible transportation between two random variables.
		Given data distribution $X$, our target is to find a bijection to transform some known distribution $Z$ to $X$, \ie, $G(Z) = X$ and $Z = G^{-1}(X)$. \hanze{{The equalities hold only if $X$ and $Z$ have same dimensions and the Jacobian of $G$ is positive definite almost everywhere.}} Then we can use a change of variable to get the density of $X$: $p_X(x) = p_Z(G^{-1}(x))\left|\nabla G^{-1}(x)\right|$. This strategy naturally leads to the MLE optimization of NF. Consider a parameterized model $G_\theta$, the log-likelihood can be written as 
		\begin{equation}
			\log p_\theta(x) = \log p_Z(G^{-1}_\theta(x)) + \log |\nabla G^{-1}_\theta(x)|.  
		\end{equation}
		
		Thus, MLE of $\theta$ can be obtained with $\nabla G$ and $G^{-1}_\theta$. This method is a free-form likelihood modeling: \HANZE{
			When $G_\theta$ is capable of transporting $Z$ to $X$, it can be successfully modelled without a specific standard distribution class.} When both $X$ and $Z$ are continuous, there exists an isomorphism $G_*$ between $Z$ and $X$
		(both of them are standard probability space \cite{rokhlin1949fundamental}), \ie,  $G^{-1}_*(X)=Z$, $G_*(Z)=X$. The only remaining issue is to approximate $G_*$ with a parameterized model $G_\theta$. When $G_*$ is well-behaved, the approximation procedure can be done smoothly.
		Nonetheless, this strategy fails
		when $X$ is scattered into several clusters. Our next example verifies this phenomenon.

		\begin{example}\label{exp:transport}
			Consider that $X$ is a 1-$d$ Gaussian mixture model with two components $\mathcal N(-\mu,\sigma^2)$, $\mathcal N(\mu,\sigma^2)$. We assume that $\mu>\sigma>0$ and  $G_*$ is the optimal transportation function from $Z$ to $X$: $G_*(Z)=X$ and $G_*^{-1}(X)=Z$. Assume that (1) $Z$ is $\mathcal N(0,1)$, then $
			|\nabla {G_{*}}(0)|=\sigma \exp\left(\frac{\mu^2}{2\sigma^2}\right)
			$.
			(2) $Z$ is a mixture of $\mathcal N(-\mu/2,\sigma^2)$ and $\mathcal N(\mu/2,\sigma^2)$, then $
			|\nabla {G_{*}}(0)|= \exp\left(\frac{3\mu^2}{8\sigma^2}\right)$.
		\end{example}
		
		\begin{example}\label{ex:latent-var}
			If we have latent variable $u$ to denote the cluster information of Example \ref{exp:transport} and we use $Z=\mathcal N(0,1)$. Then for $u=0$, $G_*(z,u=0) = \sigma z - \mu$; for $u=1$, $G_*(z,u=1) = \sigma z + \mu$. Both functions are linear and the Jacobian is a small constant, \ie, $\nabla {G_*} = \sigma$.
		\end{example}

		Example \ref{exp:transport} indicates that the Jacobian at zero can be exponentially large w.r.t. $\mu/\sigma$. We have some examples with different $\mu$ and $\sigma$ in the Appendix. For those scattered datasets, this property is fatal. Our case (2) shows that even we use a multi-clustered $Z$ in the example, without properly alignment with cluster center $\mu$, the phenomenon cannot be resolved.
		If we directly restrict $G_{\theta}$ to have  bounded  $|\nabla G_{\theta}|$  or $|\nabla^2 G_\theta|$, then the approximated transportation function is far away from  $G_*$, which is empirically verified by the examples in  
		\hanze{Figure \ref{fig:approx_fail}
		}. 
		It is shown that with the restricted Jacobian or Hessian matrix, the transportation function from $Z$ to $X$ cannot be achieved with MLE and the resulting density estimation is very poor. Hence, the idea of normalizing flow based on the plain transportation function is not reasonable in this case.
		The next example sheds light on the construction of proper transportation function to
		avoid this phenomenon. We visualize the ill-behaved transportation functions with different $\mu$ and $\sigma$ in Figure \ref{fig:disconti}.

		Example \ref{ex:latent-var} shows that the discontinuity issue can be eliminated by introducing the latent variable $u$ into the model. We will follow this motivation to build up our algorithm.

		\section{Normalizing Flow with Variational Bayes}

		In this part, we do NOT treat NF as a black box which finds MLE without further consideration about data distribution shape, which can be captured by latent representation $u$.
		In contrast, we introduce the
		latent representation parameterized by $\phi$
		to simplify $G_\theta$ and develop an algorithm based on NF to joint learn  the transportation function and latent representation.

		Intuitively,
		we first investigate data distribution properties to choose a proper type of latent representations, 
		and then use it to simplify the learning process of the transportation function.
		Similar to Figure \ref{fig:illustration}, the introduction of latent representations enables us to learn the transportation function for different modes separately.

		\vspace{.2in}
		\subsection{Design of Latent Representation}
		Since $u$ represents the properties of a data sample $x$, for reach $x$, we aims to find $u_x$ to represent the cluster (mode) information of $x$. A proper space $U$ of the latent representation  matters in the simplification of the transportation function.  We argue that the $U$ space should be capable to summarize the full cluster/mode information of data distributions. Hence, it
		is usually more fine-grained than supervised labels. 
		In the following parts, we will build up the construction and learning process of $U$ space. 
		
		Inspired by existing practical applications, we model the latent space as the following categories \cite{kingma2013auto,oord2017neural,dosovitskiy2020image}:
		(1) {\sc discrete} space: when the data are low-dimension with limited modes, such as tabular data, the  {discrete} space is sufficient for cluster information of $X$;
		(2) {\sc continuous} space: when
		the space of $x$ is close to a smooth manifold, continuous latent $u$ is a natural choice. For example, 
		the geodesic distance from starting point is a proper $u$ for Swiss roll data.
		(3) {\sc sequential} space\footnote{In a broad sense, the sequential latent space is also a discrete space, but the number of state of a sequence is usually too large to be traversed, thus we consider it as a new case.}: for images, $16\times16$ sequence is needed as a proper representation \cite{dosovitskiy2020image}. 
		Modern text representations are also based on sequences \cite{devlin2019bert, liu2019roberta, radford2019language, NEURIPS2020_1457c0d6}.
		More detailed descriptions of the properties are included in Appendix. 
		
		Specifically, we model $u$ as a random variable with prior $\pi_\beta(u)$, parameterized by $\beta$, which is required in our NF model. \hanze{
			Thus, the posterior $\pi_\theta(u|x) = p_\theta(x|u)\pi_\beta(u)/p_\theta(x)$ also be well-defined\footnote{Strictly speaking, 
				the $\pi_\theta(u|x)$ should be written as $\pi_{\theta,\beta}(u|x)$. We omit the $\beta$ in notation for simplicity without ambiguity. 
			}.
		} Then, the learning framework 
		can be established.

			\begin{table*}[t]
				\caption{Generative modeling  comparison. ($*$: With discrete latent, NVF obtains exact $p_\theta$)}
				\label{tab:bvtf}
				\vspace{0.1in}
				\centering
				\small
				\begin{sc}
					{
						\begin{tabular}{l|c|cc|ccc|c}
							\toprule
							Model&GAN&Neural-SDE&Neural-ODE&AR& VAE & NF & {\ModelName} \\
							\midrule
							free-form likelihood 	&-&\checkmark&\checkmark&\checkmark&&\checkmark&\checkmark\\
							well-behaved $|\nabla G|$ &-&&&\checkmark&\checkmark&&\checkmark\\
							exact density estimation &&&&\checkmark&&\checkmark&$*$\\
							density approximation	 &&\checkmark&\checkmark&\checkmark&\checkmark&\checkmark&\checkmark\\
							explicit generator &\checkmark&&&\checkmark&\checkmark&\checkmark&\checkmark\\
							continuous data sampling &\checkmark&\checkmark&\checkmark&&\checkmark&\checkmark&\checkmark\\
							explicit $u|x$ or $z|x$
							&&&\checkmark&\checkmark&\checkmark&\checkmark&\checkmark\\
							\bottomrule
					\end{tabular}}
			\end{sc}
				\vspace{.2in}
			\end{table*}

			\subsection{Variational Latent Posterior} We turn to present our  framework to optimize $G_\theta$ and latent representation jointly. Given latent $u$, 
			and the (prior) probability measure $\pi_\beta(u)$, we have
			\begin{align}\label{eq:likelihood}
				p_\theta(x|u) &= p_Z(G_\theta^{-1}(x,u))\left|\nabla_x
				G_\theta^{-1}(x,u)\right|, \end{align}
			\begin{align}
				p_\theta(x) &=\int p_\theta(x|u) d\pi_\beta\nonumber\\
				&=\int p_Z(G_\theta^{-1}(x,u))\left|\nabla_x  G_\theta^{-1}(x,u)\right| d\pi_\beta.   \label{eq:mle_u}
			\end{align}
			where $\pi_\beta(u)$ is the probability measure defined by $\beta$ that can be either continuous or discrete.
			
			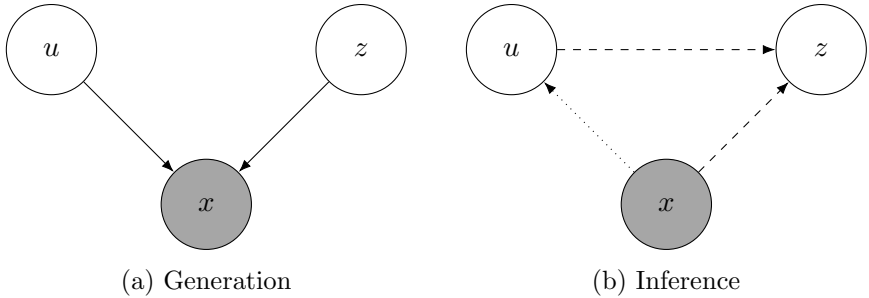
\begin{figure}
				\centering
				\begin{tabular}{cc}
					\begin{tikzpicture}
						\node[state] (1) {$u$};
						\node[state2] (2) [below right =of 1] {$x$};
						\node[state] (3) [above right =of 2] {$z$};
						
						\path (1) edge node[el,above,font=\small]  {}
						(2);
						\path (3) edge node[el,above,font=\small]{} 
						(2); 
					\end{tikzpicture}   &  
					\begin{tikzpicture}
						\node[state] (1) {$u$};
						\node[state2] (2) [below right =of 1] {$x$};
						\node[state] (3) [above right =of 2] {$z$};
						\node (4)  [left =0.1cm of 1,font=\small]{};
						\node (5)  [right =0.1cm of 3,font=\small]{};
						\path[dashed] (2) edge[bend right=0] node[el,below,font=\small] {} 
						(3);
						\path[dashed] (1) edge node[el,below,font=\small] {} 
						(3);	
						\path[dotted] (2) edge[bend left=0] node[el,below,font=\small] {}
						(1);
					\end{tikzpicture}\\
					{\small (a) Generation}&{\small (b) Inference}
				\end{tabular}

				\caption{NF with varational latent representation: sample $x$ is generated by a latent representation $u$ and noise $z$.
					Solid lines denote the NF model $G_\theta$ with latent prior $p_\beta$, such that $  p_{\theta,\beta}(x,u,z) =  p_\beta(u) p_Z(z) p_\theta(x|u,z)$, where $p_\theta(x|u,z)=1\text{ if $x=G_\theta(z,u)$}$, $p_\theta(x|u,z)=0\text{ if $x\neq G_\theta(z,u)$}$ which can also be reformulated as
					$p_\beta(u) p_Z(G_\theta^{-1}(x,u))\left|\nabla_x  G_\theta^{-1}(x,u)\right|p_\theta(z|x,u)$.
					Dotted line denotes variational approximation {{$\nu_\phi(u | x)$}} of intractable posterior $\pi_\theta(u | x)$.
					Dashed lines denote the tractable posterior $\pi_\theta(z | x,u)$.
				}
				\label{fig:dgm}
			\end{figure}
			Conventional NF methods would prefer to optimize Eq. \eqref{eq:mle_u} analytically, which is not sensible in our case: (1) For small discrete space, although it is possible to compute the exact loss, the introduction of $u$ space makes the density evaluation entail the
			traversal of $U$ space, while most $\pi_\theta(u|x)$ are near $0$ due to the clustering phenomenon; (2) When the space is large, such as continuous and sequential space, the exact optimization of Eq. \eqref{eq:mle_u} becomes intractable. Intuitively, 
			for a fixed $x$, it is wasteful to traverse the  $U$ space, since only a few of $u$ are important. Thus, we resort to a more economical way to approximate $p_\theta(x)$:
			we aim to
			obtain $u|x$ efficiently and only sample important ones.

			Variational inference provides us the possibility to find the posterior. 
			In particular, Stochastic Gradient Variational Bayes estimator enables us to obtain a variational approximator $\nu_\phi(u|x)$ of intractable posterior $\pi_\theta(u|x)$. 
			Rather than optimizing log-likelihood, we consider to optimize evidence lower bound (ELBO) to obtain $G_\theta$ and $\nu_\phi(u|x)$. 
			The full diagram of our proposed framwork is displayed in Figure \ref{fig:dgm}.

			By Jensen's inequality, 
			Proposition \ref{prop:elbo} replaces the Eq. \eqref{eq:mle_u} with a lower bound, leading to our variational formulation. 

			\begin{proposition}\label{prop:elbo}
				In our setting, the evidence lower bound (ELBO) of log-likelihood is written as 
				\begin{equation}
					\int  \log p_\theta(x|u)d\nu_\phi(u|x) -\operatorname{KL}(\nu_\phi(u|x)\Vert \pi_\beta(u)).
				\end{equation}
			\end{proposition}
			
			For efficient computation, we use Monte Carlo samples to estimate ELBO,
			\begin{equation}\label{eq:mc_elbo}
				\frac{1}{K}\sum_{k=1}^K\log p_\theta(x|u_{xk})-\operatorname{KL}(\nu_\phi(u|x)\Vert \pi_\beta(u)), 
			\end{equation}
			where $u_{xk}$ are i.i.d. samples from $\nu_{\phi}(u|x)$. 
			
			By taking Eq. \eqref{eq:likelihood} and \eqref{eq:mc_elbo} together, we can get the loss function below.
	
		\begin{proposition}\label{prop:loss}
				For each data sample $x$, we can obtain our loss function 
				\begin{align}\label{eq:loss}
					\ell(\theta,\phi, \beta;x) &=
					-
					\frac{1}{K}\sum_{k=1}^K[\log
					p_Z(G_\theta^{-1}(x,u_{xk}))
					\nonumber\\&\quad+ \log \left|\nabla_x G_\theta^{-1}(x,u)\right|]
					\nonumber\\&\quad+\operatorname{KL}(\nu_\phi(u|x)\Vert \pi_\beta(u))
				\end{align}
				where $u_{xk}$ are samples from $\nu_\phi(u|x)$.
				
			\end{proposition}

			We consider two prior parameterization strategies for different latent space (1) For discrete or continuous space, the prior $\pi_\beta(u)$ is given by uninformative distributions such as uniform distribution, Gaussian distribution, which means $\beta$ is not need to be learned; (2) For sequential space, the learning process the $\pi_\beta$ is important. With the a powerful $\pi_\beta$ class, there exists $\pi_\beta(u)$ such that $
			\pi_\beta(u) = \EE_X \nu_\phi(u|x)$, and
			the KL term can be eliminated. In such the case, the loss function can be reduced (Proposition \ref{prop:reduce}).
			
			\begin{proposition}\label{prop:reduce} If $\beta$ is learnable and $\nu_\phi$ is in the distribution class of $\pi_\beta$. Eq. \eqref{eq:loss} can be equivalently rewritten as a two separate loss functions for $\phi$, $\theta$, and $\beta$, 
				
				\noindent (1) Loss function of $\theta$ and $\phi$
				\begin{equation}	\label{eq:loss_reduce}
					\ell_1(\theta,\phi;x)=-\frac{1}{K}\sum_{k=1}^K \log p_Z(G^{-1}_\theta(x,u_{xk})) 
					- \log\left|\nabla {G^{-1}_\theta}(x)\right|;
				\end{equation}
				\\
				(2) Given $\phi$, the loss function of $\beta$ is
				\begin{align}\label{eq:mle_beta}
					\ell_2(\beta;x,\phi)= -\frac{1}{K}\sum_{i=1}^K\log\frac{d\pi_\beta(u_i)}{d\mu_u},\  u_i\sim\nu_\phi(u|x),
				\end{align}
				where we take the  Radon–Nikodym derivative of $\pi_\beta$ over the measure $\mu_u$ of $U$.
			\end{proposition}

			Thus, when $\beta$ is fixed, we minimize Eq. \eqref{eq:loss} to obtain $\hat\theta,\hat\phi$, when $\beta$ is learnable, minimization of Eq. \eqref{eq:loss_reduce} and \eqref{eq:mle_beta}
			leads to the estimator of $\hat\theta,\hat\phi,\hat\beta$.

			\vspace{.2in}
			\subsection{{\ModelName} Algorithm}\label{alg:nfvlf}
		
			To be more clear, we introduce an encoder $E_\phi(x)$ as the neural network realization of $\nu_\phi(u|x)$. For discrete $u$, the output of $E_\phi(x)$ is the probability of each state $v_i$: $E_\phi(x)=[\nu_\phi(v_1|x),\cdots,\nu_\phi(v_s|x)]^\top$. Sampling from $\nu_\phi(u|x)$ can be done by Gumble-Softmax technique \cite{jang2016categorical}. For continuous random variable, we constrain the posterior $\nu_\phi(u|x)$ can only be Gaussian, which makes output of $E_\phi(x)$ as $\mu$ and $\sigma$ for the variational posterior $\mathcal N(\mu,\sigma I)$. The sampling process is done by reparameterization trick \cite{kingma2013auto}. For sequential 
			$U$ space, $u_x = E_\phi(x)$ is a latent sequence representation, $\log \nu_\phi(u=u_x|x)\propto-\Vert u - u_x\Vert^2$,
			\HANZE{ $u_x$ is learned by vector quantization \cite{oord2017neural}}.
			$\nu_\phi(u|x)$ sample the most likely $u$. The specification of posterior sampling enable us to present full algorithm.
			\begin{algorithm*}
				\caption{{\ModelName: Normalizing Variational Flow}}
				\begin{algorithmic}[1]
					\renewcommand{\algorithmicrequire}{\textbf{Input:}}
					\renewcommand{\algorithmicensure}{\textbf{Output:}}

					\REQUIRE Data observations $\mathcal D = \{x_1,\cdots,x_n\}$; Parameter initialization $\theta$, $\phi$, and $\beta$. Learning rate $\eta$. Batch size $b$.
					\ENSURE  The learned NF $G_\theta$; variational latent posterior approximator $\nu_\phi$; the learned prior $\pi_\beta$ if it is learnable.

					\WHILE{$\theta$ and $\phi$ do not converge} 
			
					\STATE Sample data $x_{(i)} \in\mathcal D$, for $i=1,\cdots,b$. 
					
					For each $x_{(i)}$, compute $\nu_\phi(u|x_{(i)})$ and sample $u_{x_{(i)}k}$ from $\nu_\phi(u|x_{(i)})$, for $k=1,\cdots,K$.
					\STATE For each $u_{x_{(i)}k}$,  obtain the map on $Z$ space such that $z_{x_{(i)}k} = G_\theta^{-1}(x,u_{x_{(i)}k})$.
					
					\IF{$\beta$ is not learnable}
					\STATE  {\small$[\theta,\phi] \leftarrow [\theta,\phi] - \eta  b^{-1}\sum_{i=1}^b [\frac{\partial\ell(\theta,\phi,\beta;x_{(i)})}{\partial \theta},\frac{\partial\ell(\theta,\phi,\beta;x_{(i)})}{\partial \phi}]$}
					in Eq. \eqref{eq:loss}.
					\ELSE
					\STATE {\small$[\theta,\phi] \leftarrow [\theta,\phi] - \eta b^{-1}\sum_{i=1}^b [\frac{\partial\ell_1(\theta,\phi;x_{(i)})}{\partial \theta},\frac{\partial\ell_1(\theta,\phi;x_{(i)})}{\partial \phi}]$}
					in Eq. \eqref{eq:loss_reduce}.
					\STATE {\small$\beta \leftarrow \beta - \eta  b^{-1}\sum_{i=1}^b	\frac{\partial\ell_2(\beta;x_{(i)},\phi)}{\partial \beta}$} in Eq. \eqref{eq:mle_beta} .
					
					\ENDIF
					\ENDWHILE
				\end{algorithmic}
			\end{algorithm*}

			For the evaluation stage, if $u$ is discrete, we can use Eq. \eqref{eq:mle_u} to evaluate density. For continuous latent space, some approximation methods are proposed \cite{burda2015importance}, but $p_\theta(x)$ cannot be obtained efficiently, we will evaluate $p_\theta$ for this case. 
			If $u$ is sequential, we can use $E_\phi(x)$ to obtain the most likely
			$v_{x1},\cdots,v_{xK}\in U$ that are top-$K$ $\nu_\phi(u|x)$, and compute $
			\hat{p}_\theta(x) = \sum_{i=1}^K \pi_\theta(u=v_{xi}) p(x|v_{xi}).$

			\subsection{{\ModelName} v.s. Other Generative Models}
		
			In this part, we compare our {\ModelName} with other generative models. Although GAN is popular in applications \cite{goodfellow2014generative,brock2018large}, the statistical properties are missing in GAN's framework. 
			\hanze{The likelihood of is missing and we cannot even approximate the density of data samples.}   GAN-based models are based on discriminator network to approximate likelihood in feature space, whose statistical properties
			are vague \cite{goodfellow2014generative,dinh2016density}.
			Transportation based methods, such as Neural ODE/SDE and NF,
			suffer from the Jacobian issue \hanze{as described} in Section \ref{sec:back}. In particular, Neural-ODE/SDE utilizes an integral to obtain the transportation rather than an explicit function, which entails extensive computation. 
			The introduction of noise in Neural-SDE makes it unable to find the latent $z$ or $u$ reversely. 
			Autoregressive (AR) models are based on discrete/sequence modeling, which does not support continuous data sampling. 
			Compared with VAE, {\ModelName} integrates the NF framework into the likelihood modeling, which supports free-form density function and more complex data modeling, such as images. 
			\HANZE{Besides, VAE is based on the Gaussian likelihood assumption, whose modeling capacity is highly restricted.}
			Our {\ModelName} supports exact density estimation that cannot be accomplished by conventional VB frameworks. 
			Table \ref{tab:bvtf} summarizes the comparison.

			In summary, our framework shows privileges in the following scenario: 
			\hanze{
				(1) Isolated latent \emph{cluster} representations: Data distribution $X$ is constructed with both cluster center and noise nearby. The cluster representation usually captures global information of $X$, denoted as $U$. The noise part is the quotient space of $X$ by $U$, \ie, the space obtained by ``collapsing" $U$ component to zero.
				We focus on the case that the elements of $U$ are isolated. The $\nabla G$ of conventional transportation based methods (NF, Neural-ODE/SDE) would be ill-behaved.
				(2) Non-Gaussian noise space: the noise near each cluster is a complex distribution rather than Gaussian noise. The Gaussian likelihood assumption oversimplify the real practice.
				(3) Limited $G_\theta$ capacity:  The target $G_*$ with ill-behaved Jacobian/Hessian matrix cannot be achieved. The ill-behaved $\nabla G$ fails modeling.
			}

			These three conditions are automatically satisfied for most real data and parameterized function classes. We take visual datasets for example: (1) the visual datasets usually contain multiple labels and attributes. Some samples sharing the same properties belong to the same cluster. (2) The noise in image datasets varies, such as background noise, compression noise, which are not Gaussian; (3) For any parameterized function class, the capacity is limited for model size and numerical stability consideration.
			The non-smoothness and non-continuity should be avoid.
			

			
			%
			
			
			

			
			



			\section{Related Works}
			
			\subsection{Normalizing Flow} Normalizing flow (NF) is a popular and powerful framework in generative modeling. The core idea of NF is to construct a reversible generative function $G_\theta$ between a known distribution $Z$ and observation $X$, and use the change of variables formula to model the distribution of $X$. Thus the density of $X$ can be computed by $ p_Z(G^{-1}(x))\left|\nabla G^{-1}(x)\right|$. The key challenge for this framework is to obtain the inverse form of $G_\theta$ and to compute Jacobian determinant. Thus, early works \cite{dinh2014nice,kingma2018glow,dinh2016density,durkan2019neural}  target at constructing special functions of $G_\theta$ that has explicit inverse and simple Jacobian form. Some ODE-based methods \cite{chen2018neural,grathwohl2018ffjord} also consider to compute an integral to make the Jacobian more flexible.
			These algorithms make it possiple to compute the exact likelihood of data samples. Due to the explicit form of $p_\theta(x)$ for input $x$, we can perform density estimation in NF framework. Besides, the sampling procedure can also be done painlessly with the obtained  $G_\theta(z)$.
			
			Nonetheless, the transportation function might be complex. Conventional normalizing flow models \cite{dinh2014nice,dinh2016density,kingma2018glow,durkan2019neural} ignore this challenge. Interestingly, the label information of the datasets, which helps to construct better generation, can also be recognized as a latent variable \cite{brock2018large}. Our motivation originates from these ``labels"  and we extend them to general unsupervised latent variables that help transportation.

			
			\subsection{Variational Bayes} 
			Variational Bayes (VB) is a method to approximate the intractable posterior $\pi_\theta(u|x)$ of latent variables,
			which cannot be solved by conventional statistical methods such as EM algorithm. The main idea of VB is to use another parametric model $\nu_\phi(u|x)$ to approximate $\pi_\theta(u|x)$. Thus, the parameterization of $\phi$ brings approximation error regarding the capacity. 
			\hanze{Variational auto-encoder (VAE) \cite{kingma2013auto} is the most popular example that uses encoder-decoder neural network pair as the parametric model. However, the distributions of both $\pi_\theta(x|u)$ and $\nu_\phi(u|x)$ are only restricted to standard distribution classes, such as diagonal Gaussian distribution.} \hanze{VQ-VAE \cite{oord2017neural,razavi2019generating} is an extension of VAE that leverages sequential latent variable}. 
			The distribution class of $\pi_\theta(x|u)$ in VQ-VAE is still restricted.  It only considers $\pi_\theta(x|u)$ as a Gaussian distribution. 
			
			Since NF has the capacity of modeling complex densities, it is suitable for the variational approximation step \cite{rezende2015variational}. Previous work use NF for posterior approximation (VI with NF). In other words, NF was only used to model $\nu_\phi(u|x)$. On the contrary, our framework (NF with VI) does not concentrate on  complex posterior modeling but enhances NF modeling with another latent variable. It is also reasonable to consider this framework as an extension of VB that uses NF to improve the capacity of $\pi_\theta(x|u)$.
			Note that when $u$ is finite, our framework provide explicit likelihood form, which cannot be obtained by VAE formulation. We highlight that even finite latent $u$ still help NF significantly.

			\section{Experiments}
			
			We implement the variational latent encoder jointly with conditional normalizing flow.
			To justify the significance of our NVF framework, we evaluate the likelihood of our model on several datasets, such as 2d toy data, tabular data, and visual data. To provide more intuitive explanation and illustration, we provide visualization and generation results on toy and visual data, which can help readers to understand how NVF works.
			
					\begin{figure*}[t]
				\centering
				\begin{sc}
					\begin{footnotesize}
						\begin{tabular}{m{1cm}m{2.15cm}<{\centering}m{2.15cm}<{\centering}m{2.15cm}<{\centering}
								m{2.15cm}<{\centering}m{2.15cm}<{\centering}m{2.15cm}<{\centering}
							}
							Size:& $4\times4$&$4\times32$&$10\times32$
							&$4\times4$&$4\times32$&$10\times32$\\
							{\scriptsize Param.:}& 0.3K&9.0K&22K&0.3K&9.0K&22K\\
							{\scriptsize NF}&\includegraphics[width=2.15cm]{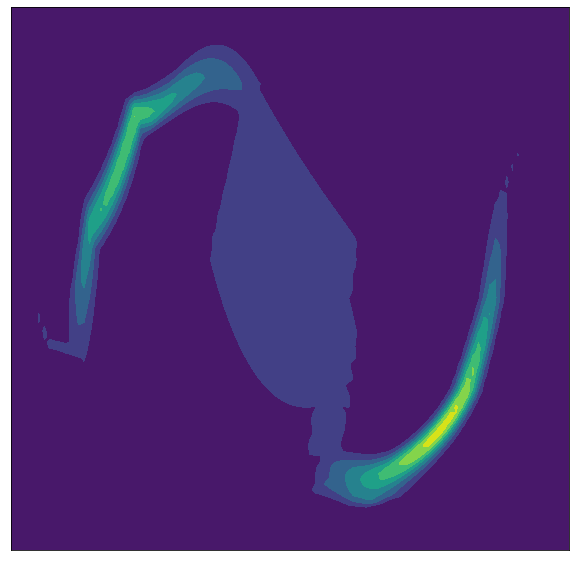}&  \includegraphics[width=2.15cm]{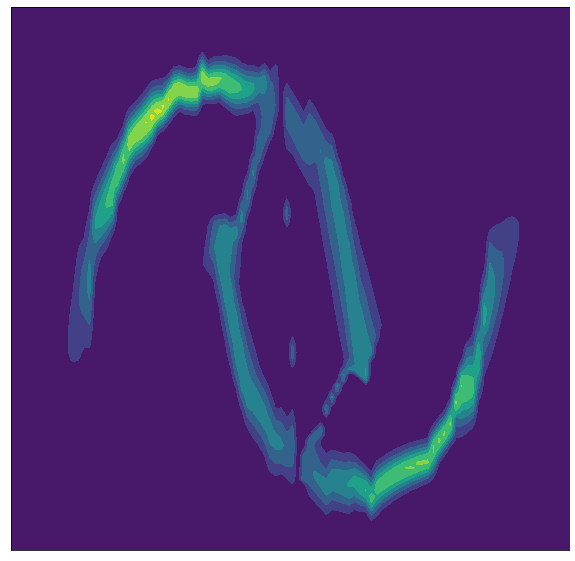}&  \includegraphics[width=2.15cm]{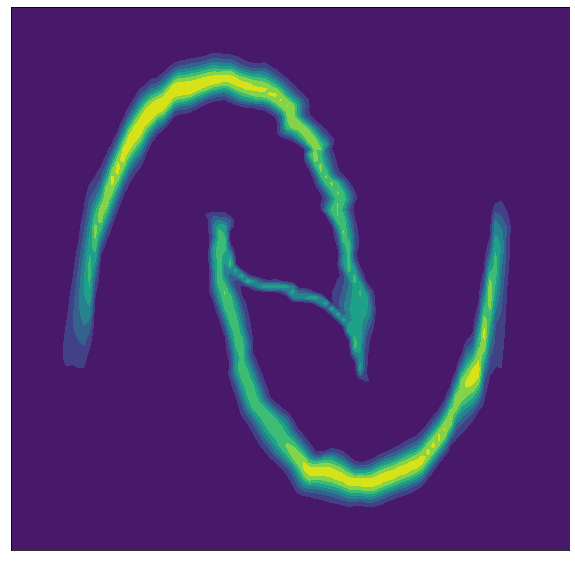}
							&\includegraphics[width=2.15cm]{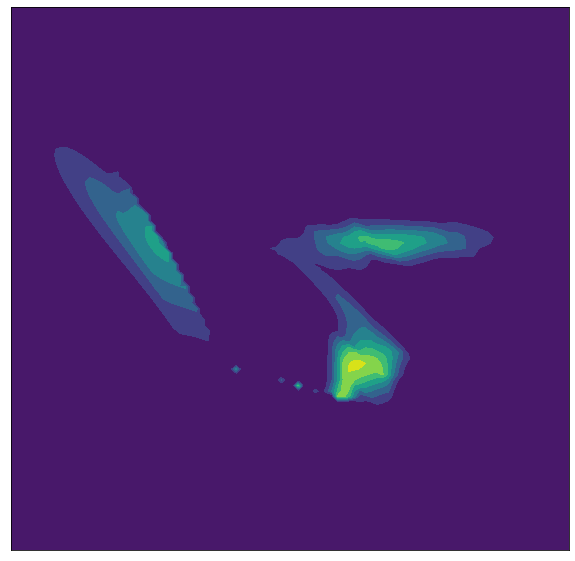}&  \includegraphics[width=2.15cm]{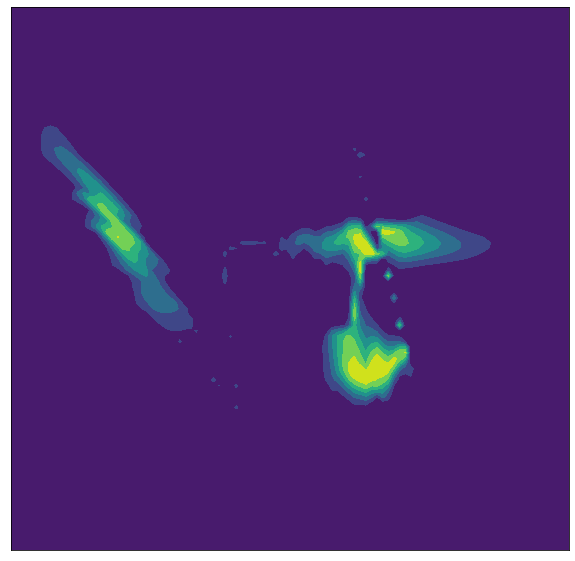}&  \includegraphics[width=2.15cm]{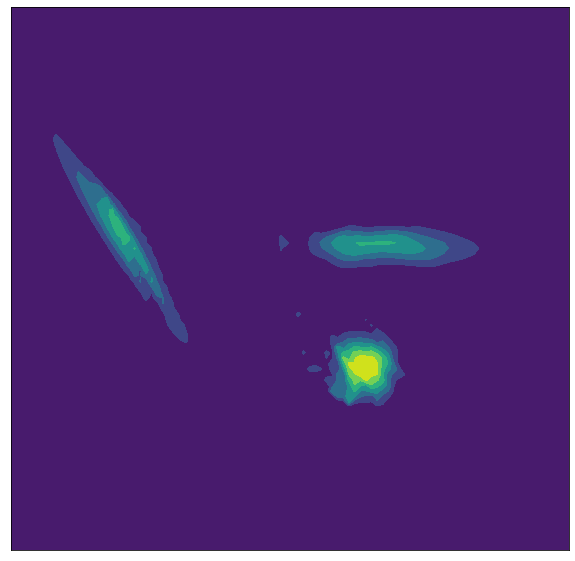}
							\\
							{\scriptsize\ModelName}&\includegraphics[width=2.15cm]{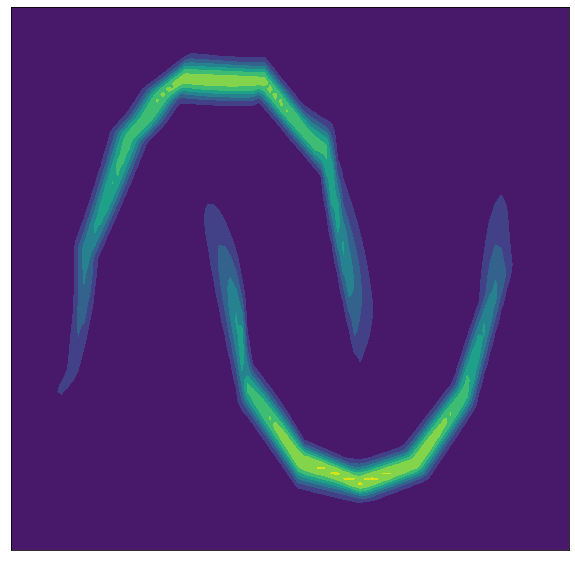} & \includegraphics[width=2.15cm]{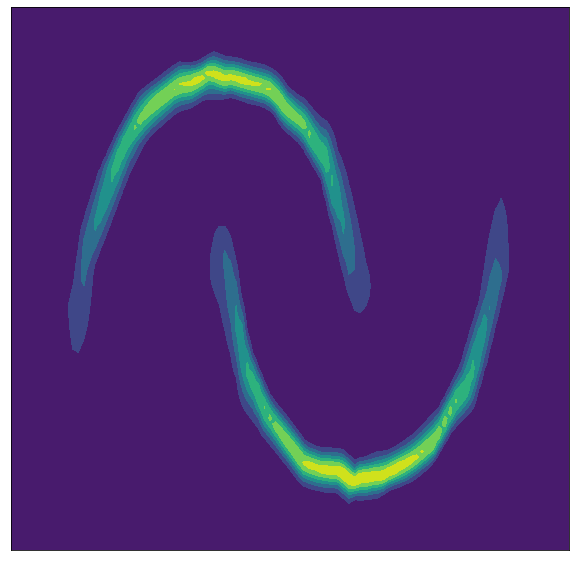} & \includegraphics[width=2.15cm]{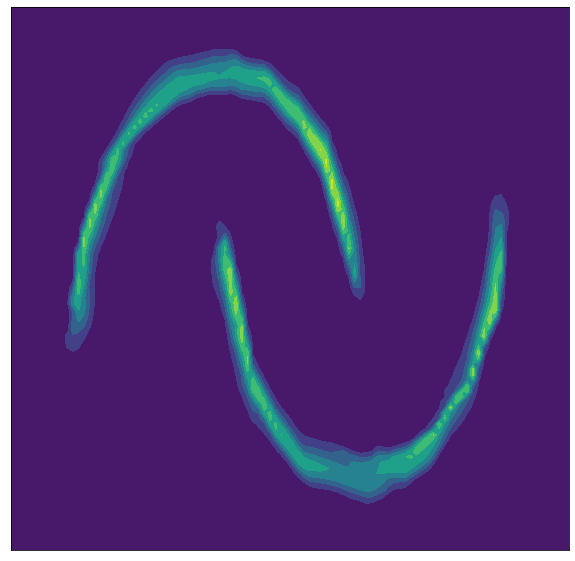}
							&\includegraphics[width=2.15cm]{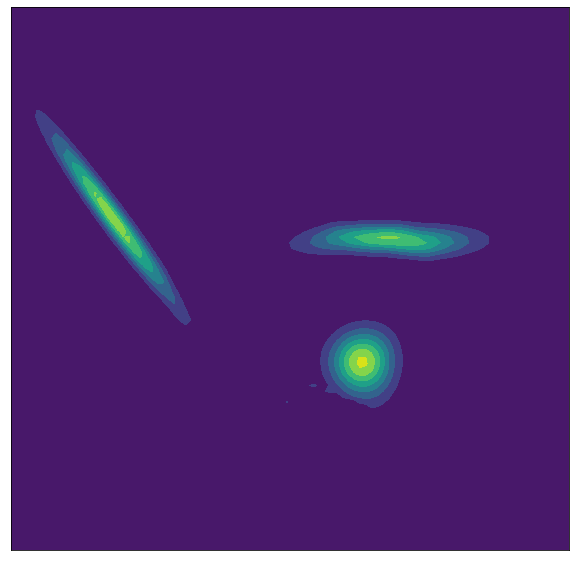} & \includegraphics[width=2.15cm]{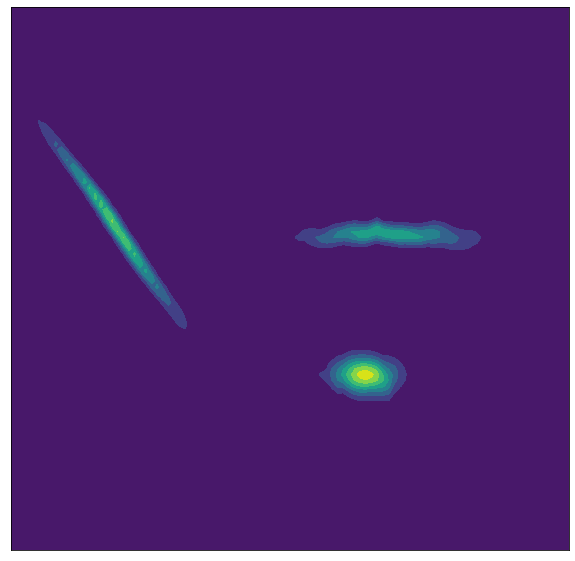} & \includegraphics[width=2.15cm]{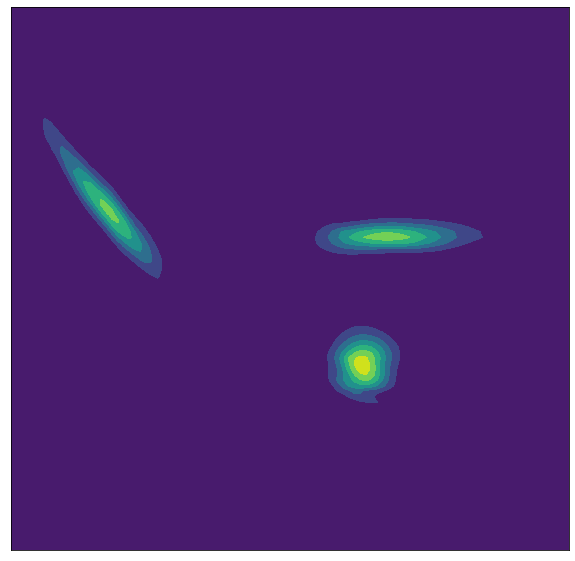}
							\\
							&Sample&Kernel density& True density&Sample&Kernel density& True density\\
							Data &\includegraphics[width=2.15cm]{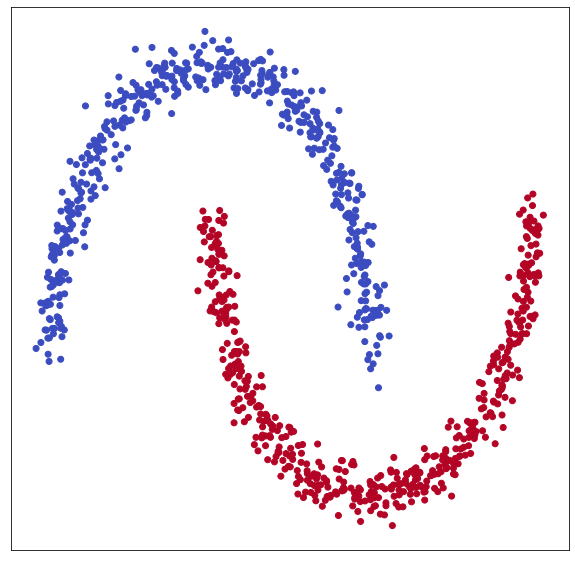}
							&\includegraphics[width=2.15cm]{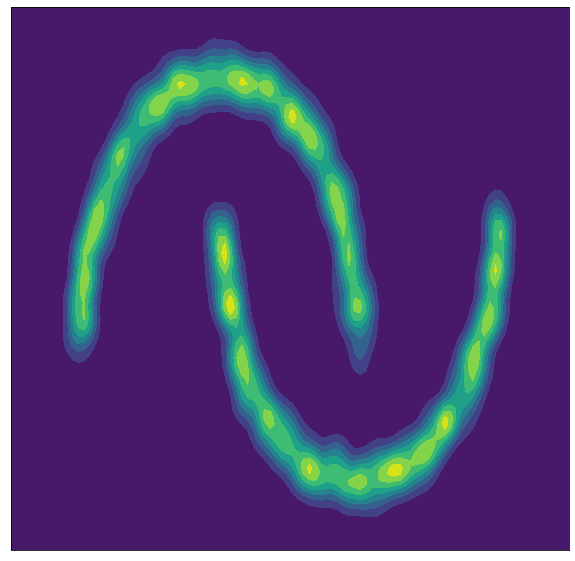}
							&\includegraphics[width=2.15cm]{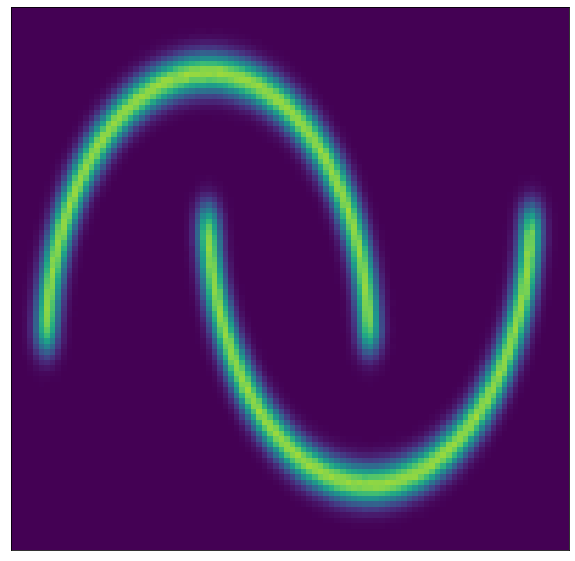}
							&
							\includegraphics[width=2.15cm]{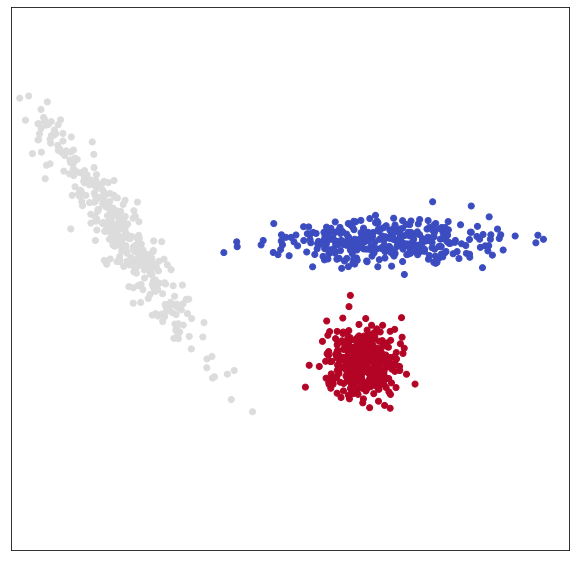}
							&\includegraphics[width=2.15cm]{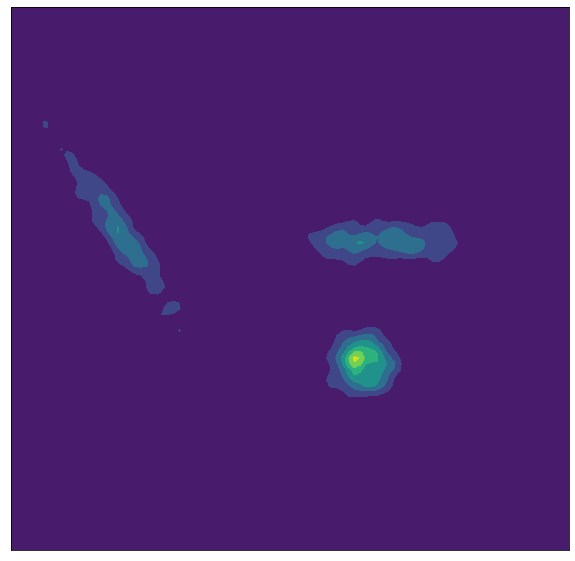}
							&\includegraphics[width=2.15cm]{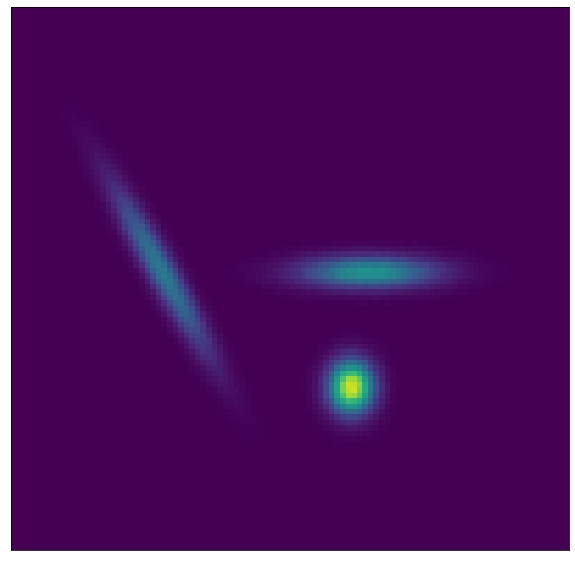}
							\\
						\end{tabular}
						
					\end{footnotesize}
				\end{sc}
				\caption{Comparison between NF  and our {\ModelName}.
					Size indicates the \emph{flow depth} $\times$ \emph{flow width}.
					{\sc Param.} indicates the parameter size. 
					For NF, we use standard Gaussian as $Z$.
					For {\ModelName}, the posterior approximator $\phi$ is with parameter $0.9K$. 
				}
				\label{fig:toy}
			\end{figure*}



			\subsection{Toy Data Illustration}
			First, we explore the representative power of our model on toy datasets, \ie, $2$-$d$ continuous data distributions. The data are generated with \emph{sklearn}. The sample size of training data is 1024. 
			Figure \ref{fig:toy} shows the comparison between NF and {\ModelName}.
			Several observations are made.
			First, with the increase of flow depths, flow widths and parameters, all three models are improved substantially and thus make a better approximation for $G_*$.
			Second, for NF models, it could not separate the modes, which is consistent with Example \ref{exp:transport}. 
			Last, our model achieves a significant improvement on fitting the moon shape and Gaussian mixture, and  the modes in NVF are separated clearly. These observations demonstrate that the endowing the normalizing flow with variational latent representation provides more powerful generative modeling capability and thus is able to fit the density better.
			

			\begin{table}
					
					\caption{Test negative log-likelihood for tabular dataset density estimation benchmarks (in nats); lower is better. (P: POWER; G: GAS; H: HEPMASS; M: MINIBOONE; B: BSDS300) The upper half denotes coupling-based flow networks; the lower half inverse/mask-autoregressive-based ones.
					} 
				
					\begin{sc}
					   
					\begin{center}
						\normalsize
						\begin{tabular}{l|ccccc} 
							\toprule
							Dataset&  P &  G &  H &  M &  B  \\
							\midrule
							{FFJORD} & -0.46 & -8.59 & 14.11 & 10.43 & -157.40  \\
							\midrule  {RealNVP} & -0.17 & -8.33 & 18.71 & 13.55 & -153.28  \\
							{Glow } & -0.17 & -8.15 & 18.92 & 11.35 & -155.07  \\
							{NSF (C)} & -0.64&-13.09&14.75&9.67&-157.54  \\
							\midrule 
							{Ours}&\textbf{-0.68}&\textbf{-13.21}&\textbf{14.53}&\textbf{7.35}&\textbf{-158.25}
							\\
							\midrule 
							\midrule 
							{MADE } & 3.08  & -3.56 & 20.98 & 15.59 & -148.85  \\
							{MAF } & -0.24 & -10.08 & 17.70 & 11.75 & -155.69  \\
							{TAN } & -0.48 & -11.19 & 15.12 & 11.01 & -157.03  \\
							{NAF } & -0.62 & -11.96 & 15.09 & 8.86 & -157.73 \\
							SOS&-0.60 &-11.99 &15.15 &8.90&-157.48 \\
							{NSF (A) } & -0.66&-13.09&14.01&9.22&-157.31  \\
							\midrule 
							{Ours}&\textbf{-0.69}&\textbf{-13.27}&\textbf{13.87}&\textbf{7.42}&\textbf{-158.26}
							\\
							\bottomrule
						\end{tabular}
					\end{center}
									\end{sc}	\label{exp:density_estimation}
				\end{table}
			
			\begin{table*}[t]
				\normalsize	
				\caption{Test negative log-likelihood for visual datasets (in bits/dim); lower is better. 
				}
				\centering
				\begin{sc}
					\begin{tabular}{l|ccccc} 
						\toprule
						Dataset& MNIST & Fashion & CIFAR10$_{\text{(5-bit)}}$ &  CIFAR10$_{\text{(8-bit)}}$ & ImageNet  \\
						\midrule  
						{RealNVP }&1.06&2.85&-&3.49&3.98\\
						{Glow } &1.05&2.95&1.67
						&3.35&3.81
						\\
						{NSF} &-&-&1.70 &3.38&3.82\\
						MARprior&0.88&-&-&3.24&3.80\\
						ResidualFlow&0.97&-&-&3.30&3.75\\
						Flow++&-&-&-&3.08&3.69\\
						SurVAE&-&-&-&3.08&3.70\\
						\midrule
						{Ours}&  \textbf{0.78} & \textbf{2.02} & \textbf{1.37}
						&
						\textbf{2.98}&\textbf{3.49}
						\\
						\bottomrule
					\end{tabular}
				\end{sc}
				\label{tab:nll_visual}
			\end{table*}

			\subsection{Latent space and likelihood
				design}
\begin{figure}
				\begin{center}
					{
						\includegraphics[scale=.7]{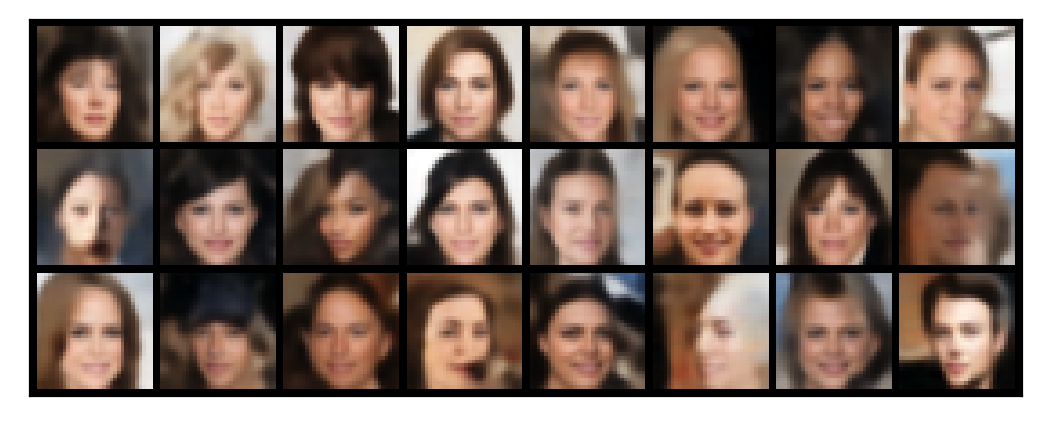}  \\
						{\footnotesize(a) VQ-VAE }\\
						\includegraphics[scale=.7]{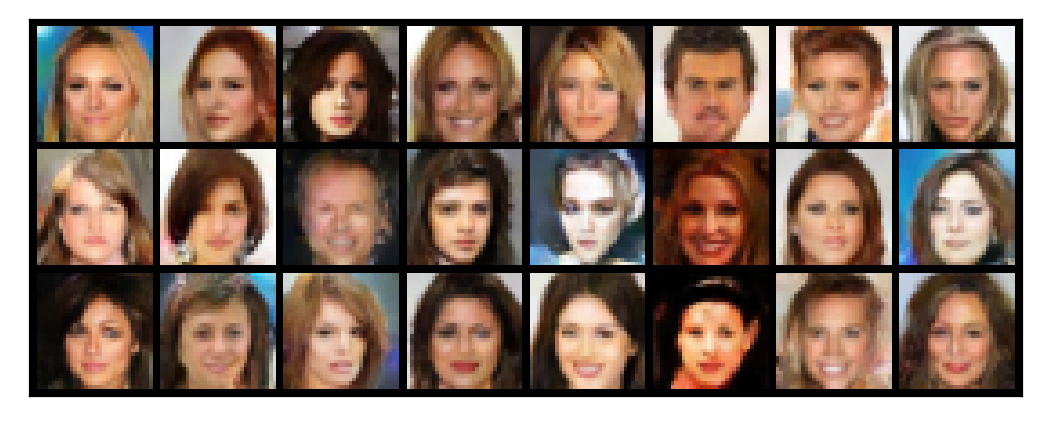} \\ 
						{\footnotesize (b) \ModelName}
					}
				\end{center}
				\caption{CelebA samples of VQ-VAE and NVF.}
				\label{fig:celeba_nvf}
			\end{figure}

			In addition, we conduct experiments on CelebA \cite{liu2015faceattributes} to discuss the effects of latent space and likelihood design from both quantitative (Table \ref{tab:fid_scores}) and qualitative results (Fig.~\ref{fig:celeba_nvf}).
			The hyper-parameter settings follow the previous section and six models are considered: VAE, VAE with discrete latent space, VQ-VAE, normalizing flow with continuous latent, normalizing flow with discrete latent and normalizing flow with sequence latent (\ie, {\ModelName}).
			First, as for the likelihood design, we observe that NF-based models outperforms VAE-based models, demonstrating the advantages of the free-form likelihood. 
			The generation samples of VAE-based method can be recognized as ``cluster centers" in our literature, since it assume the Gaussian likelihood in the model, and the nearby noise is isotropic. Our NVF does not add too much  restrictions on likelihood design, leading to better modeling.
			Furthermore, as for the latent space design, the sequence latent space performs much better than continuous and discrete latent space.
			Overall, normalizing flow with continuous latent performs the second best, verifying the good capability of normalizing flow and {\ModelName} achieves further improvement upon it, illustrating the power of sequence latent space.
			We also include the generated samples in the following sections for further comparison.

			\begin{table}[H]

					\caption{FID scores on CelebA; lower is better. Note that we display the reconstructed images for VQ-VAE due to the unsatisfied quality of generation. We compare continuous latent, discrete latent and sequence latent, separately.
					}
				
					\normalsize
					\begin{center}
						\begin{sc}
							\begin{tabular}{l|ccc} 
								\toprule
								Latent	&   continuous & discrete & sequential\\
								\midrule  
								{FID} (vae) & 47.82 & 126.27 & 31.57  \\
								{FID} (ours) & \textbf{18.85} & \textbf{68.34} & \textbf{16.23} \\
								\bottomrule
							\end{tabular}
						\end{sc}
					\end{center}
					\label{tab:fid_scores}
			\end{table}	
			
					

			\begin{figure*}[ht]
				\centering
				\begin{tabular}{cc}
					\includegraphics[scale=0.60]{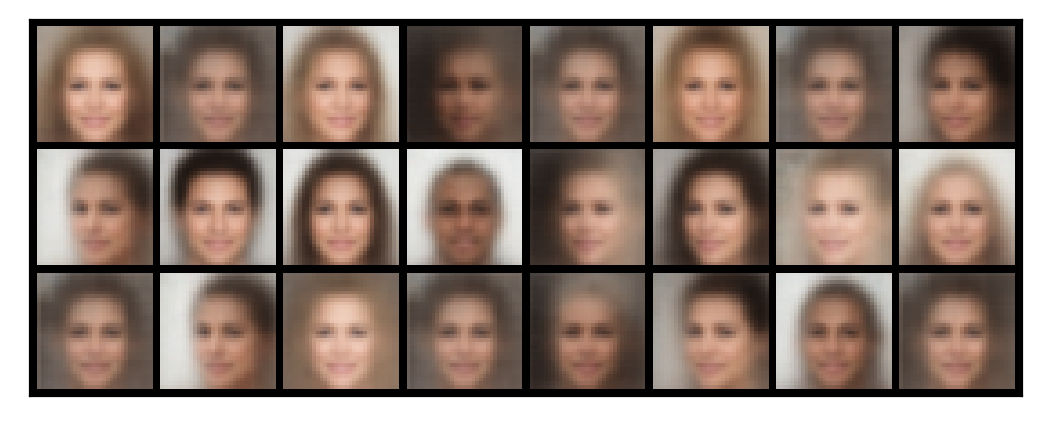} &   \includegraphics[scale=0.60]{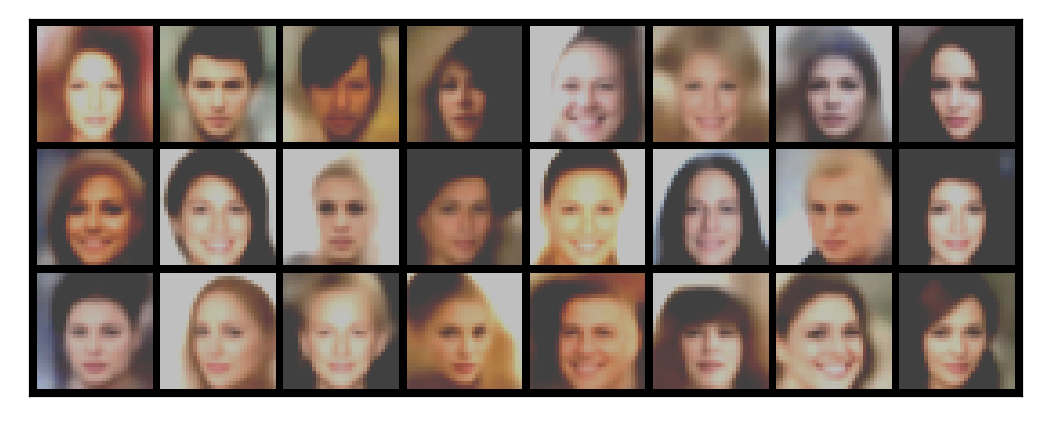}
					\\
					(a) discrete VAE& (b) VAE\\
					\includegraphics[scale=0.60]{figs/celeba_vqvae.png}&\includegraphics[scale=0.60]{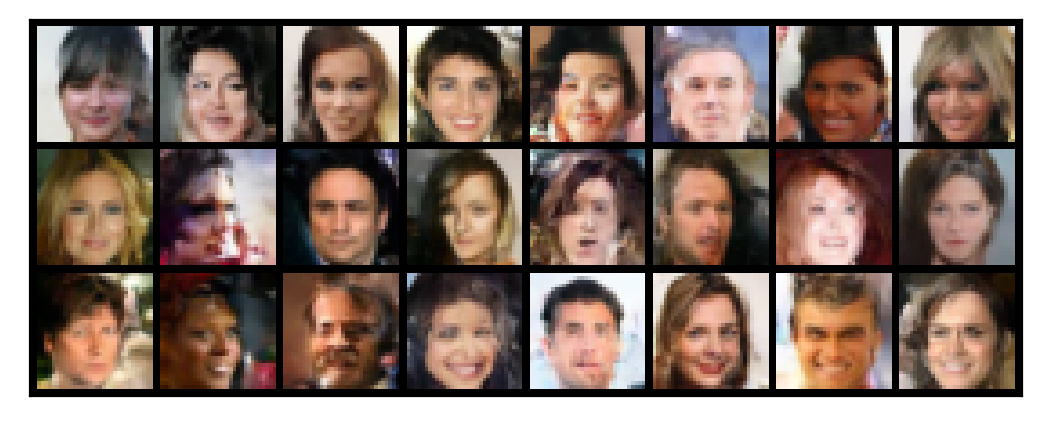}
					\\
					(c) VQ-VAE  (reconstruction) & (d) NF with continuous latent\\
					\includegraphics[scale=0.60]{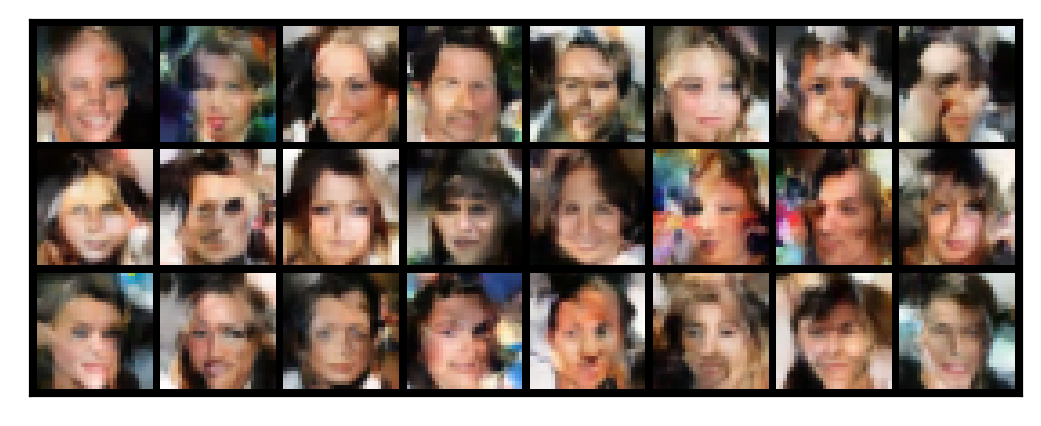}&    \includegraphics[scale=0.60]{figs/celeba_nfvlr.png}\\
					(e) NF with discrete latent  & (f) \ModelName
				\end{tabular}
				\caption{CelebA samples of different baselines.}
				\label{fig:celeba_full}
			\end{figure*}
			
			\subsection{Density estimation}
			\label{sec: density_estimation}
			We also conduct density estimation experiments on four UCI datasets with different dimensions $d$ \cite{asuncion2007uci}, namely POWER ($d=6$), GAS ($d=8$), HEPMASS ($d=21$), MINIBOONE ($d=43$); and BSDS300 dataset ($d=63$) \cite{martin2001database} with the same data pre-processing steps following \cite{papamakarios2017masked}.
			\hanze{The neural network architecture of $G_\theta$ follows NSF \cite{durkan2019neural}, with $10$ or $20$ flow depths depending on the data. The variational posterior approximator is
				a 4-layer MLP encoder with 256 neurons and ReLU activation function. The number of discrete latent states is cross-validated chosen from $\{2,5,10,20\}$.
			}
			NF has two structures in density estimation:  
			For both coupling based \cite{dinh2016density,kingma2018glow,durkan2019neural} and inverse/mask autoregressive  based \cite{papamakarios2017masked,germain2015made,kingma2016improved,jaini2019sum,oliva2018transformation,huang2018neural}  approaches, the introduction of latent representation helps significantly. 
			The results are shown in Table \ref{exp:density_estimation}.
			First, our algorithm achieves state-of-the-art performance in all density estimation benchmarks.
			Moreover, even for FFJORD which is an ODE-based algorithm with much larger capacity, our algorithm still outperform it consistently.
			Last, compared with our base architecture model NSF, our improvement is significant, especially for those datasets with higher dimensions.
			

			\subsection{Normalizing Flows on Visual Datasets}
			\label{sec: visual_datasets}
			In this section, we investigate the effectiveness of our algorithm on five real-world visual datasets:  MNIST \cite{lecun1998mnist}, Fashion MNIST \cite{xiao2017online}, CIFAR-10 \cite{krizhevsky2009learning}, CelebA \cite{liu2015faceattributes}, ImageNet \cite{deng2009imagenet}.
			Our model is trained with learning rate $10^{-4}$.
			We apply the Adam optimizer~\cite{KingmaB14} and anneal the learning rate according to a cosine schedule.
			We compare our model with Neural Spline Flow (NSF) \cite{durkan2019neural}, Glow \cite{kingma2018glow},
			MARPrior \cite{mahajan2020normalizing},
			Flow++ \cite{ho2019flow++},
			ResidualFlow \cite{chen2019residual}, 
			in terms of negative log-likelihood on the test set.
			The results are shown in Table \ref{tab:nll_visual}.
			It is observed that our model, {\ModelName}, achieves the state-of-the-art on all datasets and there are two major performance improvement.
			There is a significant jump in performance occurs between the baseline models and our {\ModelName}.
			This increase can be attributed to the power of the variational latent representation.
			Particularly, it is observed that
			our model achieves
			the best performance against
			all baseline models stuck on these datasets.
			

			
			We provide additional samples to show the importance of latent capacity and the effectiveness of NF framework in Figure \ref{fig:celeba_full}. It is clear that discrete and standard VAE is blurred, which means the capacity of Gaussian likelihood
			is insufficient to represent images. VQ-VAE can improve the performance, but the performance is still unsatisfactory (lack of details). For NF variants, due to the capacity of NF framework against Gaussian, the details can be filled. Besides, our NVF framework is the best choice to regenerate proper samples.
			



			
			\section{Conclusion}
			
			This paper introduces new framework for learning normalizing flow models, which incorporate a variational latent representation to simplify the transportation function. 
			We have shown that without the latent representation, NF fails to model data with multiple isolated modes. Our variational latent representation endows our model with strong compatibility that supports different data structures, such as images. Empirical results have shown the necessity of latent variable and the significant improvement of our framework. 
			
			\subsection*{Acknowledgments}
 The work was supported by the General
Research Fund (GRF 16310222 and GRF 16201320).

			\newpage
			
			\bibliographystyle{apalike}
			\bibliography{example_paper}
			
			

			\newpage
			\appendix

			\section{Proof}
			
            \subsection{Proof of Example \ref{exp:transport} and \ref{exp:density_estimation}}
	\begin{proof}
			Assume that the cumulative distribution function (CDF) of Standard Normal distribution is $\Phi(x)$. Then the CDF for $X$ is $$F(x) = \frac{1}{2}\left[\Phi\left(\frac{x-\mu}{\sigma}\right)+\Phi\left(\frac{x+\mu}{\sigma}\right)\right]$$
			
			For case (1), the CDF of $Z$ is $\Phi(x)$
			
			Then, by inverse transform sampling  \cite{vogel2002computational}, the optimal transportation function can be written as 
			$$
			G(z) = F^{-1}\circ\Phi(z)
			$$
			
			Thus, 
			\begin{align*}
				\frac{dG(z)}{dz} &= \frac{F^{-1}(\Phi(z))}{d \Phi(z)}\cdot \frac{d\Phi(z)}{dz}\\
				&= \frac{dF^{-1}(\Phi(z))}{d \Phi} p_Z(z) \\
				&=  \frac{p_Z(z)}{p_X(G(z))}
			\end{align*}

			When $z=0$, $\Phi(z)=\frac{1}{2}$,
			$$
			\frac{dG(z)}{dz}\bigg|_{z=0} = \frac{p_Z(0)}{p_X(0)}= \frac{1}{\frac{1}{\sigma}\exp\left(-\frac{\mu^2}{2\sigma^2}\right)}=\sigma\exp\left(\frac{\mu^2}{2\sigma^2}\right)
			$$
			
			(2) If the CDF of $Z$ is
			$$\frac{1}{2}\left[\Phi\left(\frac{x-\mu/2}{\sigma}\right)+\Phi\left(\frac{x+\mu/2}{\sigma}\right)\right]
			$$
			then similarly,
			$$
			\frac{dG(z)}{dz}\bigg|_{z=0} = \frac{p_Z(0)}{p_X(0)}=\exp\left(\frac{\mu^2}{2\sigma^2}-\frac{(\mu/2)^2}{2\sigma^2}\right) = \exp\left(\frac{3\mu^2}{8\sigma^2}\right).
			$$
			
			With latent variable, it is quite natural that $\sigma Z+\mu\sim \mathcal N(\mu,\sigma^2)$.

		\end{proof}
		\subsection{Proof of Proposition \ref{prop:elbo} and \ref{prop:loss}}
	\begin{proof}

				\begin{equation}\label{eq:elbo_proof}
\begin{aligned}
						\log p_\theta(x) &=\log \int  p_\theta(x|u) d\pi_\beta(u)\\
	&=\log \int  p_\theta(x|u) \frac{d\pi_\beta(u)}{d\nu_\phi(u)}d\nu_\phi(u)\\
	&\geq  \int  \log p_\theta(x|u) d\nu_\phi(u) + \kl(\nu_\phi\Vert\pi_\beta)
	\end{aligned}
				\end{equation}
				
				By taking
				$ p_\theta(x|u) = p_Z(G_\theta^{-1}(x,u))\left|\nabla_x G_\theta^{-1}(x,u)\right|
				$, the desired loss is obtained.
				
			\end{proof}
\subsection{Proof of Proposition \ref{prop:reduce} }
	\begin{proof}
				Since $\nu_\phi$ is in the distribution class of $\pi_\beta$, for any $\nu_\phi(u|x)$ and $X$, there exists $\beta_\phi$ such that $\pi_{\beta_\phi} = \EE_X\nu_\phi(u|x)$.
				
				Thus, by choosing $\beta_\phi$ and
				taking the expectation over $X$.
				The loss function becomes,
				\begin{align}\label{eq:loss_re}
					\EE_X\ell(\theta,\phi, \beta;x) =& 
					-\frac{1}{K}\sum_{k=1}^K \EE_X\log
					p_Z(G_\theta^{-1}(x,u_{xk}))\nonumber\\
					&-\frac{1}{K}\sum_{k=1}^K \EE_X\log \left|\nabla_x G_\theta^{-1}(x,u)\right|.
				\end{align}
				
				Thus, for each $x$, the loss function for $\phi$ and $\theta$ is equivalent to Eq. \eqref{eq:loss_reduce}.
				
				For optimization of $\beta$, the KL divergence term can also be estimated by Monte Carlo samples. Note that
				\begin{align}\label{eq:loss_kl}
					\operatorname{KL}(\nu_\phi(u|x)\Vert \pi_\beta(u)) &\propto -\int\log \frac{d\pi_\beta(u)}{d\mu_u} d\nu_\phi(u|x)
				\end{align}
				where $\mu_u$ is (1) the Lebesgue measure on $u$ if $u$ is continuous; (2) discrete measure on $u$ if it is discrete or sequential.
				
				Then the Monte Carlo estimation is 
				\begin{align}\label{eq:loss_kl2}
					-\frac{1}{K}\sum_{i=1}^K\log \frac{d\pi_\beta(u_i)}{d\mu_u},\quad u_i\sim \nu_\phi(u|x).
				\end{align}
			\end{proof}
			
			\newpage
			\section{More discussions}
			
						\subsection{Latent space $U$}

			\subsubsection{Limited discrete latent space}
			
			Discrete latent space is adopted in conventional statistics, especially for clustering analysis. 
			In particular, the idea is to divide the observation data into several groups, each group is associated with one vector: $U=\{v_1,\cdots,v_s\}$. 
			The vector $v_i$ is the representation of each group. 
			This idea is the most basic idea in representation learning that assign a feature vector for each group.
			The index of each representation in latent space can be recognized as an unsupervised label.
			However, due to the capacity of discrete latent space, it becomes less popular in current deep learning literature.
			
			
			For NF framework, the discrete latent space still has significant value. Although the NF framework has explicit likelihood, the introduction of latent variable makes the density computationally available only if $U$ is discrete (Otherwise we need to compute the integral over $u$). Meanwhile, 
			the discrete $u$ is sufficient for constructing a better transportation path in spite of the limited capacity. Thus, for tabular data, the choice of discrete $u$ is ideal to boost conventional NF.

			\subsubsection{Continuous latent space}
			The continuous latent space is the most general and widely-used representation recently. 
			Most of feature learning are based on this structure, such as the ResNet pretrained feature, latent space of VAE, etc. 
			For GAN, although we are unable to find $u|x$ explicitly, the design of $u$ is still continuous. 
			In these models, the capacity of continuous $u$ benefits the modeling process. 
			Even if the crucial latent space is discrete one, it is still a subspace of the original continuous latent space. 
			For example, the latent labels can be obtained by continuous $u$ with a linear transformation and a softmax transform. 
			
			However, from an NF view, the continuous $u$ loses an important advantage of NF models while brings powerful representation: the painless computation of $p_\theta(x)$. We need to approximate the integral over $\pi_\beta(u)$ to obtain the density form.

			\subsubsection{Sequential latent space}
			
			Another popular latent representation is the sequential ones, \ie, a sequence/tensor of latent representation. $U=\{v_1,\cdots,v_s\}^m$, where $v_i$ is the representation of each element, $m$ denote the length of sequence. The current space contains $s^m$ states, while plain discrete space only has $s$. Exponentially large state space makes the capacity totally different, so we divide it to a new category. Recently, with the development of transformers, the sequential latent space is particularly becomes a strongly competitor against continuous latent space. 
			
			Considering the NF setting, since sequential model is still discrete, the density estimation process can be done with $\nu_\phi(u|x)$ importance sampling. Meanwhile, the space of current $U$ is much larger conventional 
			discrete space, which support more complex data structures such as images, texts, etc. Notably, for this data structure, the $\pi_\beta(u)$ can be highly uneven over states, so we learn the prior from data.

					\subsection{Experimental Details}
			All of our experiments are conducted on 11GB NVIDIA 2080Ti GPUs. 
			For toy and tabular data, the encoder architecture is multilayer perceptron  (MLP). For visual datasets, the encoder architecture is based on the ResNet and the prior of latent sequence is based on a Transformer.
			We discuss the experimental details with different datasets respectively.
			
			\subsubsection*{Toy Data}
			
			On toy dataset, we use a sequence of 
			(1) Reverse Permutation;
			(2) LU Linear Transform;
			(3) Affine Coupling Transform;
			for each layer. We have compared the settings of different depth and hidden size of Affine Coupling Transform, such as [4,4], [4,16], [4,32], [10,4], [10,16], [10,32]. The posterior estimator $\nu$ is approximated by a MLP with a 4 layer $16$ hidden nodes.
			
			\subsubsection*{Tabular Data}
			
			On tabular datasets, we follow the architecture and default parameters of neural spline flows with additional context feature encoded by the posterior estimator $\nu(v|x)$.  The variational posterior approximator $\nu(v|x)$ is a 4-layer MLP encoder with hidden 256 neurons and ReLU activation function. The maximum training epoch is 1000. The model is selected by validation. Adam optimizer is used and the initial learning rate is $5\times 10^4$. 
			
			\subsubsection*{Visual Data}
			
			We first train an encoder-decoder model, where the vocabulary size is 16, embedding dimension is 16, batch size is 32, and learning rate is 1e-5.
			Then we train a Transformer model, implemented based on a 
			12-layer, 768-hidden, 12-heads GPT-2 with 512 for maximum input length and 3072 for intermediate size.
			We initialize it from the pre-trained weights of GPT-2 124M with GPT-2-simple library (\url{https://github.com/minimaxir/gpt-2-simple}).
			The Transformer model is optimized 100,000 steps with AdamW optimizer.
			The weight decay is set to 0.01 with $\beta_1 = 0.9, \beta_2 = 0.999$.
			The learning rate is 1e-4 and linearly decayed to 0 according to a cosine schedule.
			
			\subsubsection*{Time Cost Discussion}
			The time cost vs baseline in (1) Tabular data: is negligible (Sampling: $1.001\times$, Inference: $1.007\times$, $1.0\times$ indicates no extra cost). (2) CIFAR-10: is still affordable (Sampling: $3.41\times$, Inference: $1.67\times$). The computation overhead is similar to larger network.
			The training of latent prior and NF can be paralleled.
			
\end{document}